%% file: main.tex
\pdfoutput=1
\documentclass[10pt,twocolumn,letterpaper]{article}
\usepackage[margin=0.75in,columnsep=0.3125in]{geometry}
\usepackage{times}
\usepackage[T1]{fontenc}
\usepackage{amsmath,amssymb}
\usepackage{booktabs,multirow}
\usepackage{graphicx}
\usepackage[table]{xcolor}
\usepackage{xspace}
\usepackage{placeins}
\usepackage{float}
\usepackage{microtype}
\usepackage{capt-of}
\usepackage{caption}
\usepackage{tikz}
\usetikzlibrary{arrows.meta,positioning,decorations.pathreplacing,backgrounds}
\usepackage[numbers,sort&compress]{natbib}
\usepackage[colorlinks,allcolors=blue!60!black]{hyperref}
\usepackage[capitalise,nameinlink]{cleveref}
\usepackage{fontawesome5}
\definecolor{btnPage}{HTML}{1F6FD1}
\definecolor{btnCode}{HTML}{24292F}
\definecolor{btnHF}{HTML}{FFD21E}
\newcommand{\linkbutton}[4]{\href{#1}{\tikz[baseline=(b.base)]\node[rounded corners=3pt, fill=#2, text=#3,
  font=\small\sffamily\bfseries, inner xsep=6pt, inner ysep=3.5pt] (b) {#4};}}
\newcommand{\eg}{e.g.\xspace}
\newcommand{\ie}{i.e.\xspace}
\input{macros}
\title{LOCUS: Locate Once, Check Until Settled\\{\large Draft-and-Verify Visual Grounding with a Box Head and Coordinate-Token Verification}}
\newif\ifanonymous
\anonymousfalse
\ifanonymous
\author{Anonymous submission}
\else
\author{
    Muhammad Talha \\
    Independent Researcher \\
    \texttt{trizwan24@gmail.com} \\[8pt]
    \linkbutton{https://muhammad-talha-ad.github.io/LOCUS/}{btnPage}{white}{\faGlobe\ Project Page}\hspace{5pt}%
    \linkbutton{https://github.com/muhammad-talha-ad/LOCUS}{btnCode}{white}{\faGithub\ Code}\hspace{5pt}%
    \linkbutton{https://huggingface.co/MTalha2001/LOCUS-2B}{btnHF}{black}{\faCube\ LOCUS-2B}\hspace{5pt}%
    \linkbutton{https://huggingface.co/MTalha2001/LOCUS-0.8B}{btnHF}{black}{\faCube\ LOCUS-0.8B}
}
\fi
\date{}

\begin{document}
\twocolumn[{%
\renewcommand\twocolumn[1][]{#1}%
\maketitle
\vspace{-1.5em}
\begin{center}
\input{figures/teaser}
\end{center}
\vspace{1em}
}]
\input{sections/abstract}
\input{sections/intro}
\input{sections/related}
\input{sections/method}
\input{sections/setup}
\input{sections/results}
\input{sections/limitations}

{\small\raggedright

\input{main.bbl}
}

\clearpage
\appendix
\input{sections/appendix}
\end{document}

%% file: macros.tex
\newcommand{\method}{Locus\xspace}
\newcommand{\jag}{Jag\xspace}

\makeatletter
\newcommand{\N}[1]{\ifcsname num@#1\endcsname\csname num@#1\endcsname\else\textcolor{red}{\textbf{??}}\typeout{MISSING NUMBER: #1}\fi}
\makeatother
\input{generated/numbers}
\input{generated/literature_numbers}
\input{generated/refl4_numbers}

\newcommand{\accat}[1]{Acc@#1}

\newcommand{\best}[1]{\textbf{#1}}
\newcommand{\second}[1]{\underline{#1}}
\definecolor{oursc}{HTML}{2a78d6}
\newcommand{\ours}{\rowcolor{oursc!9}}
\newcommand{\up}{\,$\uparrow$}
\newcommand{\down}{\,$\downarrow$}

%% file: generated/numbers.tex
\expandafter\def\csname num@clean-jag-abs-a\endcsname{85.07}
\expandafter\def\csname num@clean-la-abs-c\endcsname{61.82}
\expandafter\def\csname num@clean-twob-head-abs-a\endcsname{88.01}
\expandafter\def\csname num@clean-twob-verified-abs-c\endcsname{54.78}
\expandafter\def\csname num@distill-distill-abs-a\endcsname{91.10}
\expandafter\def\csname num@distill-distill-abs-c\endcsname{63.56}
\expandafter\def\csname num@distill-distill-passes\endcsname{6.76}
\expandafter\def\csname num@distill-parent-abs-a\endcsname{91.10}
\expandafter\def\csname num@distill-parent-abs-c\endcsname{63.62}
\expandafter\def\csname num@distill-parent-passes\endcsname{6.85}
\expandafter\def\csname num@exact-eagerbf16-differ\endcsname{38.6}
\expandafter\def\csname num@exact-eagerbf16-gapten\endcsname{1.4}
\expandafter\def\csname num@exact-eagerbf16-n\endcsname{505}
\expandafter\def\csname num@exact-eagerbf16-other-passes\endcsname{5.66}
\expandafter\def\csname num@exact-eagerfp32-differ\endcsname{0.0}
\expandafter\def\csname num@exact-eagerfp32-gapten\endcsname{0.0}
\expandafter\def\csname num@exact-eagerfp32-n\endcsname{505}
\expandafter\def\csname num@exact-eagerfp32-other-passes\endcsname{4.39}
\expandafter\def\csname num@exact-graphedfull-differ\endcsname{35.4}
\expandafter\def\csname num@exact-graphedfull-gap\endcsname{2}
\expandafter\def\csname num@exact-graphedfull-gapten\endcsname{2.9}
\expandafter\def\csname num@exact-graphedfull-greedy-a\endcsname{92.16}
\expandafter\def\csname num@exact-graphedfull-greedy-c\endcsname{61.80}
\expandafter\def\csname num@exact-graphedfull-greedy-passes\endcsname{18.44}
\expandafter\def\csname num@exact-graphedfull-n\endcsname{10,834}
\expandafter\def\csname num@exact-graphedfull-other-a\endcsname{92.14}
\expandafter\def\csname num@exact-graphedfull-other-c\endcsname{61.89}
\expandafter\def\csname num@exact-graphedfull-other-passes\endcsname{5.41}
\expandafter\def\csname num@exact-greedybatch-differ\endcsname{48.9}
\expandafter\def\csname num@exact-greedybatch-gap\endcsname{2}
\expandafter\def\csname num@exact-greedybatch-gapten\endcsname{2.8}
\expandafter\def\csname num@exact-greedybatch-n\endcsname{505}
\expandafter\def\csname num@exact-greedygraph-differ\endcsname{51.3}
\expandafter\def\csname num@exact-greedygraph-gapten\endcsname{3.6}
\expandafter\def\csname num@exact-greedygraph-n\endcsname{505}
\expandafter\def\csname num@exact-greedyprecision-differ\endcsname{56.6}
\expandafter\def\csname num@exact-greedyprecision-gapten\endcsname{5.0}
\expandafter\def\csname num@exact-greedyprecision-n\endcsname{505}
\expandafter\def\csname num@hitoken-abs-a\endcsname{87.24}
\expandafter\def\csname num@hitoken-abs-b\endcsname{76.93}
\expandafter\def\csname num@hitoken-abs-c\endcsname{54.75}
\expandafter\def\csname num@jag-abs-a\endcsname{88.16}
\expandafter\def\csname num@jag-abs-b\endcsname{76.51}
\expandafter\def\csname num@jag-abs-c\endcsname{48.76}
\expandafter\def\csname num@jagpx-abs-a\endcsname{88.31}
\expandafter\def\csname num@jagpx-abs-c\endcsname{46.71}
\expandafter\def\csname num@la-abs-a\endcsname{88.68}
\expandafter\def\csname num@la-abs-b\endcsname{80.82}
\expandafter\def\csname num@la-abs-c\endcsname{65.55}
\expandafter\def\csname num@la-testa-a\endcsname{94.04}
\expandafter\def\csname num@lapx-abs-a\endcsname{88.26}
\expandafter\def\csname num@lapx-abs-c\endcsname{64.72}
\expandafter\def\csname num@lat-hitoken\endcsname{577.7}
\expandafter\def\csname num@lat-jag\endcsname{20.3}
\expandafter\def\csname num@lat-la\endcsname{347.5}
\expandafter\def\csname num@lat-locusvllm\endcsname{151.6}
\expandafter\def\csname num@lat-locusvllm-five\endcsname{91}
\expandafter\def\csname num@lat-locusvllm-same\endcsname{47}
\expandafter\def\csname num@lat-locusvllm-tokens\endcsname{18.3}
\expandafter\def\csname num@lat-nextchat\endcsname{436.7}
\expandafter\def\csname num@lat-qwenvllm\endcsname{297.2}
\expandafter\def\csname num@lat-qwenvllm-tokens\endcsname{41.8}
\expandafter\def\csname num@lat-qwenvllmcoords\endcsname{150.9}
\expandafter\def\csname num@lat-small-head\endcsname{25.5}
\expandafter\def\csname num@lat-twob-exact\endcsname{102.2}
\expandafter\def\csname num@lat-twob-exact-vscoords\endcsname{1.5}
\expandafter\def\csname num@lat-twob-exact-vsvllm\endcsname{2.9}
\expandafter\def\csname num@lat-twob-head\endcsname{43.5}
\expandafter\def\csname num@latq-1-exact\endcsname{94.5}
\expandafter\def\csname num@latq-1-head\endcsname{43.7}
\expandafter\def\csname num@latq-1-jag\endcsname{20.2}
\expandafter\def\csname num@latq-1-typical\endcsname{79.8}
\expandafter\def\csname num@latq-2-exact\endcsname{117.0}
\expandafter\def\csname num@latq-2-head\endcsname{44.5}
\expandafter\def\csname num@latq-2-jag\endcsname{37.4}
\expandafter\def\csname num@latq-2-typical\endcsname{94.0}
\expandafter\def\csname num@latq-4-exact\endcsname{163.8}
\expandafter\def\csname num@latq-4-head\endcsname{46.7}
\expandafter\def\csname num@latq-4-jag\endcsname{71.6}
\expandafter\def\csname num@latq-4-typical\endcsname{123.3}
\expandafter\def\csname num@latq-8-exact\endcsname{287.3}
\expandafter\def\csname num@latq-8-exact-passes\endcsname{10.3}
\expandafter\def\csname num@latq-8-exact-perq\endcsname{35.9}
\expandafter\def\csname num@latq-8-head\endcsname{53.1}
\expandafter\def\csname num@latq-8-head-perq\endcsname{6.6}
\expandafter\def\csname num@latq-8-jag\endcsname{127.3}
\expandafter\def\csname num@latq-8-jag-perq\endcsname{15.9}
\expandafter\def\csname num@latq-8-typical\endcsname{197.1}
\expandafter\def\csname num@latq-8-typical-perq\endcsname{24.6}
\expandafter\def\csname num@latq-images\endcsname{128}
\expandafter\def\csname num@leak-frac\endcsname{46.9}
\expandafter\def\csname num@leak-frac-max\endcsname{52.4}
\expandafter\def\csname num@leak-frac-min\endcsname{44.5}
\expandafter\def\csname num@mc-small-head-a-ci\endcsname{$[-0.51, +0.08]$}
\expandafter\def\csname num@mc-small-head-a-p\endcsname{0.083}
\expandafter\def\csname num@mc-small-typical-c-ci\endcsname{$[+3.86, +5.35]$}
\expandafter\def\csname num@mc-small-typical-c-p\endcsname{\ensuremath{7.2\times10^{-125}}}
\expandafter\def\csname num@mc-small-verified-c-d\endcsname{$+4.42$}
\expandafter\def\csname num@mc-smallthree-head-a-p\endcsname{0.39}
\expandafter\def\csname num@mcclean-small-head-a-ci\endcsname{$[+0.07, +0.94]$}
\expandafter\def\csname num@mcclean-small-head-a-d\endcsname{$+0.50$}
\expandafter\def\csname num@mcclean-small-head-a-p\endcsname{0.0049}
\expandafter\def\csname num@mcclean-small-head-b-ci\endcsname{$[+0.41, +1.79]$}
\expandafter\def\csname num@mcclean-small-head-b-d\endcsname{$+1.08$}
\expandafter\def\csname num@mcclean-small-head-c-ci\endcsname{$[+2.70, +4.57]$}
\expandafter\def\csname num@mcclean-small-head-c-d\endcsname{$+3.58$}
\expandafter\def\csname num@mcclean-small-typical-a-ci\endcsname{$[-0.58, +0.40]$}
\expandafter\def\csname num@mcclean-small-typical-a-d\endcsname{$-0.10$}
\expandafter\def\csname num@mcclean-small-typical-b-ci\endcsname{$[+1.36, +2.81]$}
\expandafter\def\csname num@mcclean-small-typical-b-d\endcsname{$+2.07$}
\expandafter\def\csname num@mcclean-small-typical-c-ci\endcsname{$[+9.18, +11.16]$}
\expandafter\def\csname num@mcclean-small-typical-c-d\endcsname{$+10.15$}
\expandafter\def\csname num@mcclean-small-verified-c-d\endcsname{$+10.18$}
\expandafter\def\csname num@mcclean-twob-head-a-ci\endcsname{$[+2.50, +3.46]$}
\expandafter\def\csname num@mcclean-twob-head-a-d\endcsname{$+2.95$}
\expandafter\def\csname num@mcclean-twob-head-b-ci\endcsname{$[+5.15, +6.68]$}
\expandafter\def\csname num@mcclean-twob-head-b-d\endcsname{$+5.90$}
\expandafter\def\csname num@mcclean-twob-head-c-ci\endcsname{$[+9.98, +12.13]$}
\expandafter\def\csname num@mcclean-twob-head-c-d\endcsname{$+11.03$}
\expandafter\def\csname num@mcclean-twob-verified-a-ci\endcsname{$[+1.87, +2.89]$}
\expandafter\def\csname num@mcclean-twob-verified-a-d\endcsname{$+2.35$}
\expandafter\def\csname num@mcclean-twob-verified-b-ci\endcsname{$[+5.88, +7.48]$}
\expandafter\def\csname num@mcclean-twob-verified-b-d\endcsname{$+6.65$}
\expandafter\def\csname num@mcclean-twob-verified-c-ci\endcsname{$[+14.58, +16.80]$}
\expandafter\def\csname num@mcclean-twob-verified-c-d\endcsname{$+15.66$}
\expandafter\def\csname num@nextchat-abs-a\endcsname{80.14}
\expandafter\def\csname num@nextchat-abs-b\endcsname{53.42}
\expandafter\def\csname num@nextchat-abs-c\endcsname{12.48}
\expandafter\def\csname num@selector-small-jag-a\endcsname{87.71}
\expandafter\def\csname num@selector-small-rule-a\endcsname{86.97}
\expandafter\def\csname num@selector-small-sel-a\endcsname{87.33}
\expandafter\def\csname num@selector-twob-rule-a\endcsname{89.18}
\expandafter\def\csname num@selector-twob-rule-c\endcsname{58.67}
\expandafter\def\csname num@selector-twob-sel-a\endcsname{89.44}
\expandafter\def\csname num@selector-twob-sel-c\endcsname{58.49}
\expandafter\def\csname num@small-alpha\endcsname{0.7}
\expandafter\def\csname num@small-head-abs-a\endcsname{87.97}
\expandafter\def\csname num@small-head-abs-b\endcsname{75.58}
\expandafter\def\csname num@small-head-abs-c\endcsname{47.03}
\expandafter\def\csname num@small-head-da\endcsname{$-0.19$}
\expandafter\def\csname num@small-passes\endcsname{6.68}
\expandafter\def\csname num@small-typical-abs-a\endcsname{87.17}
\expandafter\def\csname num@small-typical-abs-b\endcsname{76.14}
\expandafter\def\csname num@small-typical-abs-c\endcsname{53.29}
\expandafter\def\csname num@small-typical-dc\endcsname{$+4.52$}
\expandafter\def\csname num@small-typical-passes\endcsname{4.37}
\expandafter\def\csname num@small-verified-abs-a\endcsname{87.05}
\expandafter\def\csname num@small-verified-abs-b\endcsname{75.93}
\expandafter\def\csname num@small-verified-abs-c\endcsname{53.10}
\expandafter\def\csname num@small-verified-da\endcsname{$-1.11$}
\expandafter\def\csname num@smalljoint-head-abs-a\endcsname{87.36}
\expandafter\def\csname num@smalljoint-head-abs-c\endcsname{33.58}
\expandafter\def\csname num@smalljoint-passes\endcsname{7.98}
\expandafter\def\csname num@smalljoint-verified-abs-a\endcsname{86.36}
\expandafter\def\csname num@smalljoint-verified-abs-c\endcsname{49.79}
\expandafter\def\csname num@smalllora-head-abs-a\endcsname{88.06}
\expandafter\def\csname num@smalllora-head-abs-c\endcsname{45.72}
\expandafter\def\csname num@smalllora-passes\endcsname{7.22}
\expandafter\def\csname num@smalllora-verified-abs-a\endcsname{86.81}
\expandafter\def\csname num@smalllora-verified-abs-c\endcsname{52.52}
\expandafter\def\csname num@smallmerger-head-abs-a\endcsname{87.87}
\expandafter\def\csname num@smallmerger-head-abs-c\endcsname{46.13}
\expandafter\def\csname num@smallmerger-passes\endcsname{7.30}
\expandafter\def\csname num@smallmerger-verified-abs-a\endcsname{86.65}
\expandafter\def\csname num@smallmerger-verified-abs-c\endcsname{52.72}
\expandafter\def\csname num@smallpx-verified-abs-c\endcsname{51.25}
\expandafter\def\csname num@smallsep-head-abs-a\endcsname{87.66}
\expandafter\def\csname num@smallsep-head-abs-c\endcsname{46.64}
\expandafter\def\csname num@smallsep-passes\endcsname{6.47}
\expandafter\def\csname num@smallsep-verified-abs-a\endcsname{86.18}
\expandafter\def\csname num@smallsep-verified-abs-c\endcsname{52.25}
\expandafter\def\csname num@smallthree-head-abs-a\endcsname{88.29}
\expandafter\def\csname num@smallthree-head-abs-c\endcsname{48.86}
\expandafter\def\csname num@smallthree-head-da\endcsname{$+0.13$}
\expandafter\def\csname num@smallthree-passes\endcsname{6.82}
\expandafter\def\csname num@smallthree-verified-abs-a\endcsname{86.99}
\expandafter\def\csname num@smallthree-verified-abs-c\endcsname{52.23}
\expandafter\def\csname num@twob-a03-abs-a\endcsname{89.77}
\expandafter\def\csname num@twob-a03-abs-b\endcsname{80.53}
\expandafter\def\csname num@twob-a03-abs-c\endcsname{58.39}
\expandafter\def\csname num@twob-a03-passes\endcsname{2.64}
\expandafter\def\csname num@twob-a05-abs-a\endcsname{89.60}
\expandafter\def\csname num@twob-a05-abs-b\endcsname{80.49}
\expandafter\def\csname num@twob-a05-abs-c\endcsname{58.95}
\expandafter\def\csname num@twob-a05-passes\endcsname{3.43}
\expandafter\def\csname num@twob-a07-abs-a\endcsname{89.44}
\expandafter\def\csname num@twob-a07-abs-b\endcsname{80.37}
\expandafter\def\csname num@twob-a07-abs-c\endcsname{59.24}
\expandafter\def\csname num@twob-a07-passes\endcsname{4.04}
\expandafter\def\csname num@twob-alpha\endcsname{0.5}
\expandafter\def\csname num@twob-converged\endcsname{86.4}
\expandafter\def\csname num@twob-draft-exact2\endcsname{9.0}
\expandafter\def\csname num@twob-draft-exact4\endcsname{0.02}
\expandafter\def\csname num@twob-head-abs-a\endcsname{90.01}
\expandafter\def\csname num@twob-head-abs-b\endcsname{79.79}
\expandafter\def\csname num@twob-head-abs-c\endcsname{54.87}
\expandafter\def\csname num@twob-passes\endcsname{5.60}
\expandafter\def\csname num@twob-selected-abs-a\endcsname{89.60}
\expandafter\def\csname num@twob-selected-abs-c\endcsname{59.07}
\expandafter\def\csname num@twob-tau\endcsname{0.3}
\expandafter\def\csname num@twob-tokens\endcsname{18.46}
\expandafter\def\csname num@twob-typical-abs-a\endcsname{89.60}
\expandafter\def\csname num@twob-typical-abs-b\endcsname{80.49}
\expandafter\def\csname num@twob-typical-abs-c\endcsname{58.95}
\expandafter\def\csname num@twob-typical-passes\endcsname{3.43}
\expandafter\def\csname num@twob-verified-abs-a\endcsname{89.30}
\expandafter\def\csname num@twob-verified-abs-b\endcsname{80.31}
\expandafter\def\csname num@twob-verified-abs-c\endcsname{59.13}
\expandafter\def\csname num@twob-verified-dc\endcsname{$+10.37$}
\expandafter\def\csname num@twobdual-head-abs-a\endcsname{89.73}
\expandafter\def\csname num@twobdual-head-abs-c\endcsname{43.09}
\expandafter\def\csname num@twobdual-passes\endcsname{7.07}
\expandafter\def\csname num@twobdual-verified-abs-a\endcsname{88.74}
\expandafter\def\csname num@twobdual-verified-abs-c\endcsname{60.29}
\expandafter\def\csname num@twobjoint-head-abs-a\endcsname{88.70}
\expandafter\def\csname num@twobjoint-head-abs-c\endcsname{37.57}
\expandafter\def\csname num@twobjoint-verified-abs-a\endcsname{88.05}
\expandafter\def\csname num@twobjoint-verified-abs-c\endcsname{60.29}
\expandafter\def\csname num@twoblora-head-abs-a\endcsname{89.81}
\expandafter\def\csname num@twoblora-head-abs-c\endcsname{44.80}
\expandafter\def\csname num@twoblora-passes\endcsname{6.86}
\expandafter\def\csname num@twoblora-verified-abs-a\endcsname{88.78}
\expandafter\def\csname num@twoblora-verified-abs-c\endcsname{60.51}
\expandafter\def\csname num@twobpx-verified-abs-c\endcsname{58.10}
\expandafter\def\csname num@zs-jag-a\endcsname{91.50}
\expandafter\def\csname num@zs-jag-b\endcsname{78.81}
\expandafter\def\csname num@zs-jag-c\endcsname{48.83}
\expandafter\def\csname num@zs-qwenvl-a\endcsname{88.53}
\expandafter\def\csname num@zs-qwenvl-b\endcsname{79.52}
\expandafter\def\csname num@zs-qwenvl-c\endcsname{59.56}
\expandafter\def\csname num@zs-qwenvlfour-a\endcsname{90.60}
\expandafter\def\csname num@zs-qwenvlfour-b\endcsname{81.44}
\expandafter\def\csname num@zs-qwenvlfour-c\endcsname{61.92}
\expandafter\def\csname num@zs-smallhead-a\endcsname{91.46}
\expandafter\def\csname num@zs-smallhead-b\endcsname{78.57}
\expandafter\def\csname num@zs-smallhead-c\endcsname{47.46}
\expandafter\def\csname num@zs-smallverified-a\endcsname{90.30}
\expandafter\def\csname num@zs-smallverified-b\endcsname{78.84}
\expandafter\def\csname num@zs-smallverified-c\endcsname{53.97}
\expandafter\def\csname num@zs-twobhead-a\endcsname{93.09}
\expandafter\def\csname num@zs-twobhead-b\endcsname{82.58}
\expandafter\def\csname num@zs-twobhead-c\endcsname{55.81}
\expandafter\def\csname num@zs-twobverified-a\endcsname{91.99}
\expandafter\def\csname num@zs-twobverified-b\endcsname{82.86}
\expandafter\def\csname num@zs-twobverified-c\endcsname{60.54}

%% file: generated/literature_numbers.tex
\expandafter\def\csname num@lit-ferret13b-test\endcsname{85.3}
\expandafter\def\csname num@lit-florence2l-test\endcsname{91.1}
\expandafter\def\csname num@lit-groundingdino-test\endcsname{86.7}
\expandafter\def\csname num@lit-jag08b-test\endcsname{88.2}
\expandafter\def\csname num@lit-locateanything3b-test\endcsname{87.7}
\expandafter\def\csname num@lit-locus2b-head-test\endcsname{90.1}
\expandafter\def\csname num@lit-qwen25vl3b-test\endcsname{84.7}
\expandafter\def\csname num@lit-qwen25vl7b-test\endcsname{86.2}
\expandafter\def\csname num@lit-shikra7b-test\endcsname{82.5}

%% file: generated/refl4_numbers.tex
\expandafter\def\csname num@refl4-hitoken-o365bigimg-a\endcsname{1.9}
\expandafter\def\csname num@refl4-hitoken-o365smallimg-a\endcsname{40.6}
\expandafter\def\csname num@refl4-jag-coco-c\endcsname{39.97}
\expandafter\def\csname num@refl4-jag-o365-a\endcsname{35.84}
\expandafter\def\csname num@refl4-jag-o365q1-a\endcsname{0.0}
\expandafter\def\csname num@refl4-jag-o365q4-a\endcsname{81.5}
\expandafter\def\csname num@refl4-la-coco-c\endcsname{64.38}
\expandafter\def\csname num@refl4-la-o365-a\endcsname{77.95}
\expandafter\def\csname num@refl4-la-o365q1-a\endcsname{68.1}
\expandafter\def\csname num@refl4-n\endcsname{31,921}
\expandafter\def\csname num@refl4-ncoco\endcsname{7,804}
\expandafter\def\csname num@refl4-ntrain\endcsname{12,085}
\expandafter\def\csname num@refl4-ntrainrows\endcsname{321,327}
\expandafter\def\csname num@refl4-o365-nbig\endcsname{1,939}
\expandafter\def\csname num@refl4-o365-q1-max\endcsname{1.3}
\expandafter\def\csname num@refl4-o365-q4-min\endcsname{5.8}
\expandafter\def\csname num@refl4-ono\endcsname{12,032}
\expandafter\def\csname num@refl4-qwenvl-coco-c\endcsname{57.07}
\expandafter\def\csname num@refl4-qwenvl-o365-a\endcsname{74.45}
\expandafter\def\csname num@refl4-qwenvl-o365q1-a\endcsname{61.8}
\expandafter\def\csname num@refl4-qwenvl-o365q4-a\endcsname{86.8}
\expandafter\def\csname num@refl4-relarea-o365\endcsname{2.4}
\expandafter\def\csname num@refl4-relarea-train\endcsname{15.6}
\expandafter\def\csname num@refl4-relarea-train-p1\endcsname{4.9}
\expandafter\def\csname num@refl4-twobhead-o365q1-a\endcsname{0.0}
\expandafter\def\csname num@refl4-twobverified-coco-c\endcsname{57.09}
\expandafter\def\csname num@refl4-twobverified-o365-a\endcsname{47.81}
\expandafter\def\csname num@refl4-twobverified-o365q1-a\endcsname{6.2}
\expandafter\def\csname num@refl4-twobverified-o365q4-a\endcsname{84.8}
\expandafter\def\csname num@refl4-words\endcsname{24}

%% file: figures/teaser.tex
\definecolor{tA}{HTML}{1baf7a}
\definecolor{tB}{HTML}{2a78d6}
\definecolor{tE}{HTML}{eb6834}
\definecolor{tN}{HTML}{8a8986}
\centering
\begin{tikzpicture}[
  font=\footnotesize, >=Stealth,
  panel/.style={rounded corners=5pt, draw=#1!45, fill=#1!7, line width=0.6pt},
  rpanel/.style={rounded corners=5pt, draw=black!18, fill=black!2, line width=0.6pt},
  ptitle/.style={font=\normalsize\bfseries, text=#1!70!black},
  rtitle/.style={font=\normalsize\bfseries, text=black!80},
  body/.style={font=\scriptsize, align=left, text=black!80},
  ct/.style={draw=white, line width=0.4pt, font=\ttfamily\tiny, minimum width=2.4mm, minimum height=2.9mm, inner sep=0pt}]
\draw[panel=tA] (0, 0) rectangle (5.75, -3.85);
\node[ptitle=tA, anchor=north west] at (0.15, -0.1) {1.\ Draft};
\node[inner sep=0, anchor=north west] (d) at (0.2, -0.65) {\includegraphics[width=2.55cm]{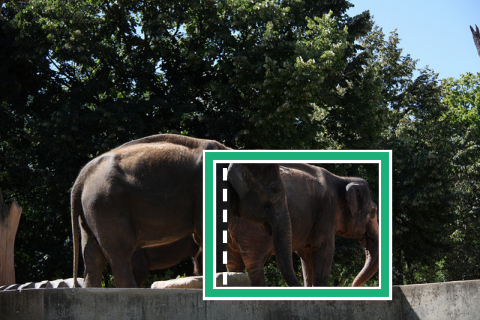}};
\node[font=\scriptsize\itshape, anchor=north] at (d.south) {``right elephant''};
\node[body, anchor=north west, text width=2.75cm] at (2.9, -0.65) {A small head on one \texttt{[BOX]} token predicts
  the whole box in one pass over the cached image.\\[3pt] The draft is often loose; here it extends onto the other
  elephant (IoU 0.87).};
\draw[->, line width=1.2pt, tA!80!black] (5.85, -0.85) -- (11.9, -0.85)
  node[midway, above=1pt, font=\scriptsize, text=black!75] {render the box as coordinate tokens};
\begin{scope}[shift={(6.25, -1.45)}]
  \node[anchor=west, font=\scriptsize\bfseries, text=black!80] at (0, 0) {Sequential steps for this example};
  \node[anchor=west, font=\scriptsize, text=black!70] at (0, -0.42) {greedy decoding};
  \foreach \i in {0,...,18} \fill[tN!55] (0.05+0.215*\i, -0.62) rectangle ++(0.17, -0.3);
  \node[anchor=west, font=\scriptsize\bfseries, text=black!70] at (4.15, -0.77) {19};
  \node[anchor=west, font=\scriptsize, text=black!70] at (0, -1.2) {\method: 1 head pass + 3 verifier passes};
  \fill[tA] (0.05, -1.4) rectangle ++(0.17, -0.3);
  \foreach \i in {1,2,3} \fill[tB] (0.05+0.215*\i, -1.4) rectangle ++(0.17, -0.3);
  \node[anchor=west, font=\scriptsize\bfseries, text=tB!80!black] at (0.95, -1.55) {4 steps, identical answer};
  \node[anchor=west, font=\scriptsize\itshape, text=black!60, text width=5.2cm] at (0, -2.15)
    {on average \N{twob-passes} passes instead of \N{twob-tokens} steps};
\end{scope}
\draw[panel=tB] (12.0, 0) rectangle (17.74, -3.85);
\node[ptitle=tB, anchor=north west] at (12.2, -0.1) {2.\ Verify};
\begin{scope}[shift={(12.38, -0.95)}]
  \node[font=\tiny, anchor=west, text=black!60] at (-0.1, 0.3) {draft as tokens};
  \foreach [count=\i from 0] \t in {4,3,4,{,},\textvisiblespace,4,8,6,{,},\textvisiblespace,8,0,1,{,},\textvisiblespace,9,1,4,]}
    \node[ct, fill=tA!28] at (0.25*\i, 0) {\t};
  \node[font=\tiny, anchor=west, text=black!60] at (-0.1, -0.4) {pass 1: keep the agreed prefix, correct the next token};
  \node[ct, fill=tA!70] at (0, -0.7) {4};
  \node[ct, fill=tE!80, text=white] at (0.25, -0.7) {\textbf{7}};
  \foreach \i in {2,...,18} \node[ct, fill=black!8, text=black!35] at (0.25*\i, -0.7) {?};
  \node[font=\tiny, anchor=west, text=black!60] at (-0.1, -1.1) {fixed point $=$ the model's greedy answer};
  \foreach [count=\i from 0] \t in {4,7,4,{,},\textvisiblespace,4,8,8,{,},\textvisiblespace,8,0,2,{,},\textvisiblespace,9,1,0,]}
    \node[ct, fill=tB!28] at (0.25*\i, -1.4) {\t};
  \foreach \i/\t in {1/7, 7/8, 12/2, 17/0} \node[ct, fill=tB!75, text=white] at (0.25*\i, -1.4) {\textbf{\t}};
\end{scope}
\node[inner sep=0, anchor=north east] (f) at (17.65, -2.5) {\includegraphics[width=1.25cm]{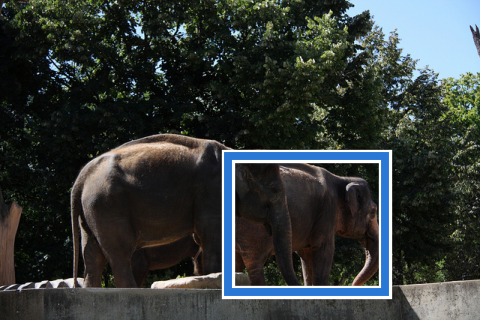}};
\node[body, anchor=north west, text width=3.95cm] at (12.15, -2.5) {The backbone checks all tokens in parallel and
  corrects the first disagreement, until the draft is stable: its greedy box (IoU 0.96).};
\def\w{4.28}
\def\g{0.2}
\def\top{-4.05}
\foreach \k/\t in {0/Acc@0.9, 1/Acc@0.5, 2/Latency, 3/Train--test overlap} {
  \draw[rpanel] (\k*\w+\k*\g, \top) rectangle ++(\w, -4.3);
  \node[rtitle, anchor=north west] at (\k*\w+\k*\g+0.15, \top-0.1) {\t};
}
\foreach \k/\f in {0/acc90, 1/acc50, 2/speed}
  \node[inner sep=0, anchor=north] (c\k) at (\k*\w+\k*\g+0.5*\w, \top-0.7) {\includegraphics{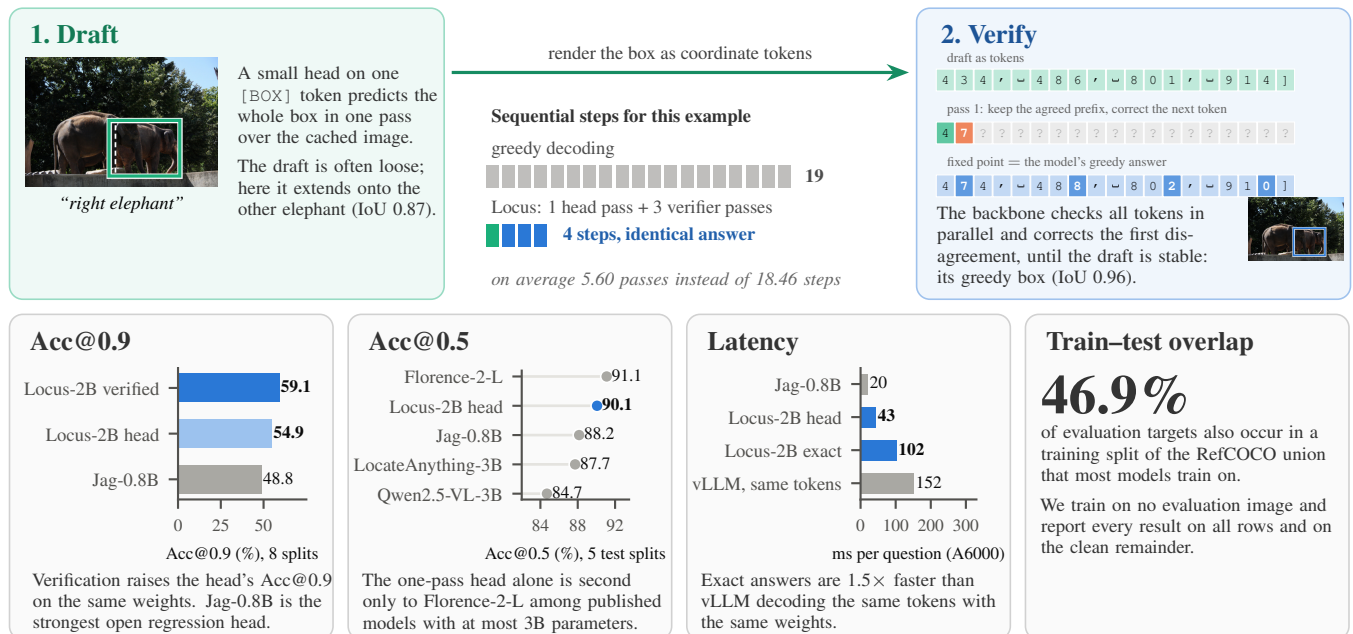}};
\node[body, anchor=north west, text width=4.0cm] at (c0.south west) {Verification raises the head's Acc@0.9 on the
  same weights. \jag-0.8B is the strongest open regression head.};
\node[body, anchor=north west, text width=4.0cm] at (c1.south west) {The one-pass head alone is second only to
  Florence-2-L among published models with at most 3B parameters.};
\node[body, anchor=north west, text width=4.0cm] at (c2.south west) {Exact answers are
  \N{lat-twob-exact-vscoords}$\times$ faster than vLLM decoding the same tokens with the same weights.};
\node[anchor=north west, font=\huge\bfseries, text=black!80, inner sep=0] (h) at (3*\w+3*\g+0.2, \top-0.8)
  {\N{leak-frac}\%};
\node[body, anchor=north west, text width=3.85cm, inner sep=0] (h1) at ([yshift=-4pt]h.south west) {of evaluation
  targets also occur in a training split of the RefCOCO union that most models train on.};
\node[body, anchor=north west, text width=3.85cm, inner sep=0] at ([yshift=-6pt]h1.south west) {We train on no evaluation image and report
  every result on all rows and on the clean remainder.};
\end{tikzpicture}
\captionof{figure}{Overview of \method. Top: a box head drafts the box in one pass (green), and the backbone verifies
the draft as coordinate tokens in a few parallel passes (blue), returning its own greedy answer. Bottom: accuracy at
IoU 0.9 and 0.5, latency, and the overlap between RefCOCO-family training and evaluation targets. All numbers are
measured in our harness unless marked as reported; latency is on one RTX A6000 at batch 1.}
\label{fig:teaser}

%% file: sections/abstract.tex
\begin{abstract}
Vision-language models (VLMs) ground a referring expression in one of two ways. They decode the box as coordinate
tokens, which is precise but takes one sequential step per token, or they regress it from a hidden state with a small
head, which takes one step but loses accuracy at high IoU. The two fail differently: a head usually finds the right
object but places its edges loosely, while coordinate tokens place edges precisely. We therefore use the regression
head as a \emph{speculative drafter} for the model's own coordinate tokens. \method writes the head's box as
coordinate tokens and checks them with the same backbone by Jacobi iteration over a shared image prefix. The fixed
point is the model's greedy coordinate-token answer, reached in about \N{twob-passes} sequential passes on average
instead of \N{twob-tokens} decoding steps; head and verifier are one set of weights, trained jointly. On the 8
RefCOCO, RefCOCO+ and RefCOCOg splits, verification raises \method-2B's \accat{0.9} from \N{twob-head-abs-c} for its
own head to \N{twob-verified-abs-c}, and the one-pass head is ahead at \accat{0.5} of every published model with at
most 3B parameters except Florence-2-L. Against \jag, the strongest open regression-head model, at its own size and
training budget, \method ties at \accat{0.5} and, with typical acceptance, gains \N{small-typical-dc} points at
\accat{0.9}, with paired significance tests. Exact answers
come \N{lat-twob-exact-vscoords}$\times$ faster than vLLM greedy-decoding the same coordinate tokens with the same
weights; a pure regression head remains faster still.
\end{abstract}

%% file: sections/intro.tex
\section{Introduction}
\label{sec:intro}

Referring-expression grounding asks a model to return the box of the object a phrase describes, a building block of
visual question answering, interactive image editing and agents that act on what they see. Modern VLMs answer by writing the
box as text: Qwen-VL and its successors~\cite{bai2023qwenvl,qwen3vl}, Shikra~\cite{chen2023shikra},
Kosmos-2~\cite{peng2023kosmos2}, Ferret~\cite{you2023ferret} and Florence-2~\cite{xiao2023florence2} decode four
coordinates token by token. Coordinate tokens are precise, because each one is conditioned on the image and on the
coordinates already written, but a box costs about \N{twob-tokens} sequential decoding steps.

Regression heads remove that cost but lose precision. \jag~\cite{zhu2026jag} attaches a 0.53M-parameter MLP to the
last input token of a 0.8B VLM and predicts all four coordinates in one forward pass. It matches coordinate-token
decoding at \accat{0.5} at a fraction of the latency, but its own matched-backbone ablation shows the price: decoding
remains clearly better at \accat{0.75} and \accat{0.9}. The two outputs fail in different ways. Without any grounding
fine-tuning, Qwen3-VL-2B's greedy coordinate tokens trail \jag at \accat{0.5} (\N{zs-qwenvl-a} vs \N{zs-jag-a}) yet
lead it at \accat{0.9} (\N{zs-qwenvl-c} vs \N{zs-jag-c}, \cref{tab:zeroshot}). The same holds within one model that has
both outputs on the same weights (\method-2B: \N{twob-head-abs-a} vs \N{twob-verified-abs-a} at \accat{0.5},
\N{twob-head-abs-c} vs \N{twob-verified-abs-c} at \accat{0.9}, head vs tokens). The head more often finds the right
object, and the tokens place its edges more exactly.

Speculative decoding~\cite{leviathan2023speculative,chen2023accelerating} addresses the same trade-off in text
generation: a cheap drafter proposes tokens, the exact model verifies them in parallel, and the output is the exact
model's. \method
(\cref{fig:teaser}) applies it to grounding, with a box head on the same backbone as the drafter in place of a smaller
language model. The head's box is written in the verifier's own coordinate tokens, and the backbone checks every token
at once by Jacobi iteration~\cite{santilli2023accelerating} over a cached image prefix, keeping the prefix it agrees
with and correcting the rest. Because the head usually finds the right object, a draft is typically wrong only in a
few digits, and a few passes reach the fixed point, which is the backbone's greedy coordinate-token answer. The image
is encoded once, and every pass is a short branch of a few tokens.

\paragraph{Contributions.}
\begin{itemize}
\item \textbf{Draft-and-verify grounding.} A box head drafts the box, and the backbone's own coordinate-token decoding
verifies it. With exact acceptance the output equals the greedy coordinate-token answer (identical on every one of
\N{exact-eagerfp32-n} questions in fp32, and within the rounding that already makes bf16 greedy decoding
irreproducible), in \N{twob-passes} sequential passes on average. Typical acceptance~\cite{cai2024medusa} trades
exactness for fewer passes (\cref{sec:method:verify}).
\item \textbf{Joint training.} Head and verifier share one set of weights and are trained together, on one image
prefill per step with isolated question branches; two separately trained models are no more accurate and need two
prefills (\cref{sec:results:ablations}). A dual mode keeps answers to other questions about the image
exactly the base model's (\cref{sec:method:train}).
\item \textbf{High-IoU accuracy.} On the same weights, verification raises \method-2B's \accat{0.9} from
\N{twob-head-abs-c} to \N{twob-verified-abs-c} over the 8 RefCOCO/+/g splits (\cref{sec:results:main}). At the size
and training budget of \jag, the strongest open regression head, \method ties it at \accat{0.5} and, with typical
acceptance, gains \N{small-typical-dc} points at \accat{0.9} (paired McNemar tests and an image-level bootstrap,
\cref{sec:results:equal}). The one-pass 2B head alone is ahead at \accat{0.5} of every published model with
at most 3B parameters except Florence-2-L (\cref{sec:results:literature}).
\item \textbf{Latency.} \method's exact answers are \N{lat-twob-exact-vscoords}$\times$
faster than vLLM decoding the same model's coordinate tokens (\N{lat-twob-exact-vsvllm}$\times$ faster than the base
model's full JSON answer); a pure regression head is faster still (\cref{sec:results:latency}).
\end{itemize}

We also make the evaluation stricter. The RefCOCO, RefCOCO+ and RefCOCOg splits are drawn per dataset, and
\N{leak-frac}\% of their evaluation rows ask about an object that is in another dataset's training split; models
trained on the union of the three, as is common, have seen those objects. We train on no evaluation image and report
results both on all rows and on the clean rows. There, \method-0.8B's head, which ties \jag-0.8B at \accat{0.5} on all
rows, is ahead (\N{mcclean-small-head-a-d}, \cref{sec:results:clean}).

\paragraph{Scope.} Our main comparison is with regression heads. A larger model trained on more data,
LocateAnything-3B~\cite{wang2026locateanything}, remains ahead of \method at high IoU (\cref{sec:results:main}). Like \jag, \method is
trained on the RefCOCO family alone, and on Ref-L4's Objects365 images, whose targets are smaller than any in RefCOCO,
both fall far below the zero-shot base model (\cref{tab:refl4}).

%% file: sections/related.tex
\section{Related Work}
\label{sec:related}
\input{figures/architecture}

\paragraph{Coordinate-token grounding.} Most VLMs ground an expression by writing its box as text, as plain numbers in
Shikra~\cite{chen2023shikra} and the Qwen-VL family~\cite{bai2023qwenvl,qwen3vl}, location tokens in
Kosmos-2~\cite{peng2023kosmos2}, Florence-2~\cite{xiao2023florence2} and PaliGemma~\cite{beyer2024paligemma}, and
hybrid region features in Ferret~\cite{you2023ferret}. Recent work changes the vocabulary or the order:
Hi-Token~\cite{zhu2026hitoken} gives each axis and digit position its own tokens, and
LocateAnything~\cite{wang2026locateanything} emits a whole box per step. All of them still produce the answer by
decoding. In \method such a model is the verifier, and \method reaches its greedy answer in fewer sequential steps.

\paragraph{Regression and decoder heads on VLMs.} \jag~\cite{zhu2026jag} regresses the box from the last input token
with a small MLP, and is the closest work: our drafter follows its design. Earlier systems route a special token's
embedding to a separate decoder: NExT-Chat~\cite{zhang2023nextchat} for boxes, LISA~\cite{lai2023lisa} for masks, and
VisionLLM~v2~\cite{wu2024visionllmv2} for many task decoders. Our head also predicts an IoU-aware confidence, as dense
detectors do~\cite{wu2020iouaware,zhang2021varifocalnet,li2020gfl}. All of these systems return the head's output as
the answer; \method returns it only after the coordinate-token model agrees.

\paragraph{Speculative and parallel decoding.} Speculative decoding~\cite{leviathan2023speculative,chen2023accelerating}
drafts tokens with a small model and verifies them with the target model in one parallel pass, preserving the target
distribution. Medusa~\cite{cai2024medusa} drafts with extra heads on the target model and introduces typical
acceptance; Jacobi decoding~\cite{santilli2023accelerating} and lookahead decoding~\cite{fu2024lookahead} iterate a
whole draft to its fixed point. Their drafts are tokens, from a smaller language model, from extra token heads or
from the target itself. Ours is a regression head trained on ground-truth boxes, whose continuous output is rendered into the verifier's tokens,
and it drafts the entire answer at once.

\paragraph{Shared-prefix computation.} Prefix caching and shared-prefix attention~\cite{zheng2023sglang,juravsky2024hydragen}
reuse one prompt's keys and values across requests. We use the same idea inside one request: the image is encoded
once and every question, draft and verification pass is a branch over its cache.

%% file: figures/architecture.tex
\definecolor{cvis}{HTML}{eda100}
\definecolor{ctxt}{HTML}{8a8986}
\definecolor{cbox}{HTML}{1baf7a}
\definecolor{cver}{HTML}{2a78d6}
\definecolor{cfix}{HTML}{eb6834}
\begin{figure*}[t]
\centering
\begin{tikzpicture}[
  font=\footnotesize, >=Stealth, line width=0.5pt,
  blk/.style={draw=black!70, rounded corners=2pt, align=center, inner sep=2.5pt},
  tk/.style={draw=white, line width=0.6pt, minimum width=3.4mm, minimum height=3.4mm, inner sep=0pt},
  ct/.style={draw=white, line width=0.5pt, font=\ttfamily\scriptsize, minimum width=3.1mm, minimum height=3.6mm,
             inner sep=0pt},
  lab/.style={font=\scriptsize, text=black!65, align=center},
  arr/.style={->, draw=black!70}]
\begin{scope}[on background layer]
  \draw[rounded corners=5pt, draw=black!18, fill=black!2, line width=0.6pt] (-0.5, 3.55) rectangle (14.9, -1.2);
\end{scope}
\node[anchor=west, font=\small\bfseries] at (-0.3, 3.2) {(a) Architecture};
\node[inner sep=0, anchor=west] (img) at (-0.3, 1.05) {\includegraphics[width=1.95cm]{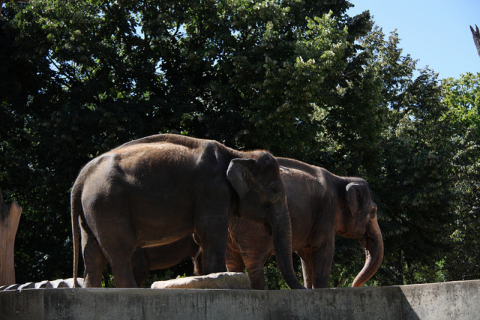}};
\node[lab, anchor=north, align=center] at (img.south) {image $I$,\\$q$ = ``right elephant''};
\node[blk, fill=black!8, minimum height=6mm, align=center] (vit) at (2.8, 1.05) {vision encoder\\{\scriptsize(frozen)}};
\node[blk, fill=cver!14, minimum height=5mm] (mrg) at (2.8, 0.0) {merger {\scriptsize(adapted)}};
\draw[arr] (img.east) -- (vit.west);
\draw[arr] (vit) -- (mrg);
\def\ty{0.0}
\foreach \i in {0,...,5} \node[tk, fill=cvis!70] (v\i) at (4.05+0.36*\i, \ty) {};
\foreach \i in {0,1} \node[tk, fill=ctxt!45] (p\i) at (6.3+0.36*\i, \ty) {};
\foreach \i in {0,1} \node[tk, fill=cbox!35] (q\i) at (7.45+0.36*\i, \ty) {};
\node[tk, fill=cbox, font=\tiny\bfseries, text=white] (bx) at (8.17, \ty) {B};
\foreach \i in {0,1,2} \node[tk, fill=cver!30] (r\i) at (8.85+0.36*\i, \ty) {};
\foreach \i in {0,...,4} \node[tk, fill=cver!75] (c\i) at (9.93+0.36*\i, \ty) {};
\draw[arr] (mrg.east) -- (v0.west);
\draw[decorate, decoration={brace, amplitude=3pt, mirror}] (3.9,-0.27) -- (6.62,-0.27)
  node[midway, below=3pt, lab] {image tokens + prefix $p$:\\prefilled once, cached};
\draw[decorate, decoration={brace, amplitude=3pt, mirror}] (7.3,-0.27) -- (8.33,-0.27)
  node[midway, below=3pt, lab] {box branch:\\$q$ + \texttt{[BOX]}};
\draw[decorate, decoration={brace, amplitude=3pt, mirror}] (8.7,-0.27) -- (11.6,-0.27)
  node[midway, below=3pt, lab] {verifier branch: grounding\\prompt for $q$ + coordinate tokens};
\node[blk, fill=cver!7, minimum width=78mm, minimum height=7mm, anchor=south west] (llm) at (3.8, 0.4)
  {language model {\scriptsize(fine-tuned; the base copy answers other questions)}};
\foreach \n in {v0,v5,p1,q1,bx,r1,c2} \draw[arr, draw=black!35] (\n.north) -- (\n.north |- llm.south);
\node[blk, fill=cbox!18] (mlp) at (7.55, 1.85) {box head: LN $\rightarrow$ MLP $\rightarrow\sigma$\\[-1pt]{\scriptsize\color{black!65}train: $\ell_1$ + GIoU + IoU-aware BCE}};
\draw[arr] (bx.north |- llm.north) to[out=90, in=-90] (mlp.south);
\node[lab, anchor=south, text=black!80, align=center] (bout) at ([yshift=2mm]mlp.north) {box $\hat b$\\+ confidence $c$};
\draw[arr] (mlp) -- (bout);
\node[blk, fill=cver!18] (lmh) at (10.45, 1.85) {LM head\\[-1pt]{\scriptsize\color{black!65}train: token CE}};
\draw[arr] (c2.north |- llm.north) to[out=90, in=-90] (lmh.south);
\node[blk, fill=cver!6, font=\ttfamily\scriptsize, anchor=south] (tok) at ([yshift=2mm]lmh.north) {474, 488, 802, 910]};
\draw[arr] (lmh) -- (tok);
\begin{scope}[shift={(12.75, 2.75)}]
  \def\s{0.25}
  \node[anchor=south, lab] at (4*\s, 0.02) {branch attention mask};
  \foreach \r/\c/\col in {0/0/cvis, 1/0/cvis, 2/0/cvis, 3/0/cbox, 4/0/cbox, 5/0/cver, 6/0/cver, 7/0/cver,
                          1/1/cvis, 2/1/cvis, 3/1/cbox, 4/1/cbox, 5/1/cver, 6/1/cver, 7/1/cver,
                          2/2/cvis, 3/2/cbox, 4/2/cbox, 5/2/cver, 6/2/cver, 7/2/cver,
                          3/3/cbox, 4/3/cbox, 4/4/cbox,
                          5/5/cver, 6/5/cver, 7/5/cver, 6/6/cver, 7/6/cver, 7/7/cver}
    \fill[\col!60] (\c*\s, -\r*\s) rectangle ++(\s-0.03, -\s+0.03);
  \draw[black!40] (0, 0) rectangle (8*\s, -8*\s);
  \node[lab, rotate=90] at (-0.18, -1.5*\s) {prefix};
  \node[lab, rotate=90] at (-0.18, -4*\s) {box};
  \node[lab, rotate=90] at (-0.18, -6.5*\s) {verify};
  \node[lab, anchor=north, text width=27mm] at (4*\s, -8*\s-0.05)
    {each branch sees the prefix and itself, never another branch};
\end{scope}
\begin{scope}[shift={(0, -1.95)}]
\begin{scope}[on background layer]
  \draw[rounded corners=5pt, draw=black!18, fill=black!2, line width=0.6pt] (-0.5, 0.35) rectangle (14.9, -3.5);
\end{scope}
\node[anchor=west, font=\small\bfseries] at (-0.3, 0.0) {(b) Draft-and-verify inference};
\def\x{4.75}   %
\def\dx{0.31}  %
\def\lx{4.55}  %
\node[inner sep=0, anchor=north west] (bd) at (-0.2, -0.4) {\includegraphics[width=2.3cm]{figures/example_draft.png}};
\node[lab, anchor=north, align=center] at (bd.south) {\textbf{1. draft} from the head\\(1 pass, IoU 0.87)};
\node[lab, anchor=east, align=right] at (\lx, -0.7) {\textbf{2. write as tokens}};
\foreach [count=\i from 0] \t in {4,3,4,{,},\textvisiblespace,4,8,6,{,},\textvisiblespace,8,0,1,{,},\textvisiblespace,9,1,4,]}
  \node[ct, fill=cbox!28] (d\i) at (\x+\dx*\i, -0.7) {\t};
\node[lab, anchor=east, align=right] at (\lx, -1.25) {\textbf{3. verify}, pass 1};
\node[ct, fill=cbox!70] at (\x, -1.25) {4};
\node[ct, fill=cfix!80, text=white] at (\x+\dx, -1.25) {\textbf{7}};
\foreach \i in {2,...,18} \node[ct, fill=black!8, text=black!35] at (\x+\dx*\i, -1.25) {?};
\node[lab, anchor=east, align=right] at (\lx, -1.75) {passes 2--3};
\node[lab, anchor=west] at (\x-0.15, -1.75) {\textit{same rule on the new draft, every position in parallel}};
\node[lab, anchor=east, align=right] at (\lx, -2.25) {\textbf{4. fixed point}};
\foreach [count=\i from 0] \t in {4,7,4,{,},\textvisiblespace,4,8,8,{,},\textvisiblespace,8,0,2,{,},\textvisiblespace,9,1,0,]}
  \node[ct, fill=cver!28] at (\x+\dx*\i, -2.25) {\t};
\foreach \i/\t in {1/7, 7/8, 12/2, 17/0}  %
  \node[ct, fill=cver!75, text=white] at (\x+\dx*\i, -2.25) {\textbf{\t}};
\node[anchor=west, lab, inner sep=0] at (\x-0.15, -2.75) {%
  \tikz[baseline=-0.6ex]{\node[ct, fill=cbox!70]{};} accepted: the model agrees\qquad
  \tikz[baseline=-0.6ex]{\node[ct, fill=cfix!80]{};} corrected: replaced by the model's own token};
\node[anchor=west, lab, inner sep=0] at (\x-0.15, -3.15) {%
  \tikz[baseline=-0.6ex]{\node[ct, fill=black!8]{};} re-predicted, checked in the next pass\qquad
  \tikz[baseline=-0.6ex]{\node[ct, fill=cver!75]{};} final tokens that differ from the draft};
\node[inner sep=0, anchor=north west] (bf) at (10.95, -0.4) {\includegraphics[width=2.3cm]{figures/example_final.png}};
\node[lab, anchor=north, align=center] at (bf.south) {\textbf{5. verified box}\\(greedy answer, IoU 0.96)};
\draw[arr] (\x+\dx*18.5, -2.25) -- ++(0.4, 0) |- ([yshift=-3mm]bf.west) ;
\end{scope}
\end{tikzpicture}
\caption{Architecture and inference of \method, illustrated with \method-2B on the RefCOCO val expression ``right
elephant'' (dashed white: ground truth). (a) The image and a fixed prefix are prefilled once. Each question then runs as
a short branch over the cached prefix (attention mask on the right), so all box and verifier branches of an image share
one prefill. The head reads the final state of the learned \texttt{[BOX]} token; the verifier branch is the backbone's
own grounding prompt, read by its LM head. Head and verifier are trained jointly. (b) The head's box is rendered as the
coordinate tokens the model would write, and the model checks all of them in one parallel pass, keeping the longest
prefix it agrees with (green) and correcting the next token (orange). Repeating this until the draft stops changing
yields the model's greedy answer, here in 3 passes instead of 19 decoding steps. Blue tokens are those that
verification changed.}
\label{fig:overview}
\end{figure*}

%% file: sections/method.tex
\section{Method}
\label{sec:method}

\method has three parts that share one image prefill: a box head that drafts (\cref{sec:method:head}), a verifier that
checks the draft with the backbone's own coordinate tokens (\cref{sec:method:verify}), and a training recipe that
learns both in one set of weights (\cref{sec:method:train}). \Cref{fig:overview} shows the
whole pipeline on one question.

\subsection{Shared prefill and question branches}
\label{sec:method:prefill}

Verification runs several passes per question, and a request may ask several questions about one image; encoding the
image again for each would cost more than the passes save. Let $I$ be an image and $q_1,\dots,q_Q$ the expressions to
ground in it. The backbone encodes $[I; p]$ once, where $p$
is a fixed instruction prefix, and caches the resulting keys and values (or recurrent states, for hybrid backbones).
Each question then runs as a \emph{branch}: a short token sequence that attends to the cached prefix and to itself but
not to other branches. A branch is therefore exactly what a separate request with the same prefix would compute, while
the image costs one encoder and one prefill per image rather than per question. Box, verifier and decision questions
are all branches of this kind.

\subsection{Box head (drafter)}
\label{sec:method:head}

A box branch is the expression followed by a box-query token with a learned embedding. Following \jag, a small MLP
reads the final-layer state $h$ of that token and predicts a normalised box
$\hat b = \sigma(\mathrm{MLP}(\mathrm{LN}(h)))$ in centre-size form, together with a confidence logit $c$. The head is
trained with $\mathcal{L}_\text{box} = 5\,\lVert \hat b - b\rVert_1 + 2\,(1-\mathrm{GIoU}(\hat b, b))$ and an
IoU-aware binary cross-entropy that targets $\mathbf{1}[\mathrm{IoU}(\hat b, b)\ge 0.5]$ for $c$. One branch pass
yields the draft; this is the ``head'' answer in our tables.

\subsection{Coordinate-token verification}
\label{sec:method:verify}

The verifier is the adapted backbone (\cref{sec:method:train}) answering the grounding prompt the base model was
pretrained on, \eg ``Locate $q$ in the image and output its bbox coordinates in JSON format'', after a teacher-forced
JSON preamble. The target is this model's greedy coordinate-token answer. It is a token
sequence $y^\ast = (y^\ast_1,\dots,y^\ast_T)$ spelling four integers in $[0,1000]$ and a closing bracket.

\paragraph{Rendering.} The draft box is rounded to the same integer grid and tokenised exactly as the verifier would
write it, giving a draft sequence $d^{(0)}$.

\paragraph{Jacobi passes.} Pass $k$ runs one verifier branch on $d^{(k)}$ and reads the argmax $g_t$ at every position
in parallel, each conditioned on the draft before it. Let $n$ be the length of the longest prefix with
$g_t = d^{(k)}_t$. Those $n$ tokens and the next prediction $g_{n+1}$ are exact greedy tokens, because each was
predicted from an exact prefix. The next draft keeps them and replaces the rest by the remaining predictions,
$d^{(k+1)} = (d^{(k)}_{1:n}, g_{n+1}, g_{n+2},\dots)$. Iteration stops when the draft no longer changes.

\paragraph{Exactness.} Every pass extends the verified prefix by at least one token, so the iteration reaches a fixed
point within $T$ passes, and a fixed point satisfies $g_t = d_t$ for all $t$, \ie it is the greedy decode. In practice
we cap the passes at 8 and finish any unconverged row by greedy decoding from its verified prefix, so the answer is
always $y^\ast$. A perfect draft costs one pass; a draft wrong only in its last digits costs two or three.
The greedy answer is \N{twob-tokens} tokens long on average, so plain greedy decoding takes that many sequential
steps; exact verification takes \N{twob-passes} at 2B (\N{small-passes} at 0.8B). \N{twob-converged}\% of questions
reach the fixed point within 8 passes, and the greedy steps of the rest are included in these counts.

\paragraph{Exactness in fp32.} The argument assumes each position's argmax does not depend on how the computation is
batched. In fp32 that holds for our implementation: on \N{exact-eagerfp32-n} RefCOCO val questions the verified box
equals the greedy box every time (\N{exact-eagerfp32-differ}\% differ, \cref{tab:exactness}).

\paragraph{Exactness in bf16.} In bf16 the assumption fails for greedy decoding itself. Rerunning plain greedy decoding with two images per batch instead of one changes
\N{exact-greedybatch-differ}\% of the boxes, and switching to CUDA graphs changes \N{exact-greedygraph-differ}\%,
almost always by \N{exact-greedybatch-gap} units of the 0--1000 grid on one coordinate, where two digits are nearly
tied. Verification differs from greedy decoding at the same rate (\N{exact-graphedfull-differ}\% of all
\N{exact-graphedfull-n} val questions on our accuracy path, median \N{exact-graphedfull-gap} units) with the same
accuracy (\accat{0.5} \N{exact-graphedfull-other-a} vs.\ \N{exact-graphedfull-greedy-a}, \accat{0.9}
\N{exact-graphedfull-other-c} vs.\ \N{exact-graphedfull-greedy-c}). ``Exact'' therefore means exact up to the
rounding that makes bf16 greedy decoding irreproducible in the first place. Rounding also costs passes: tied digits
can flip between passes, and the same \N{exact-eagerfp32-n} questions converge in \N{exact-eagerfp32-other-passes}
passes in fp32 against \N{exact-eagerbf16-other-passes} in bf16.

\paragraph{Typical acceptance.} With $\alpha<1$, a draft token is also accepted when the verifier assigns it at least
$\alpha$ times the probability of its argmax~\cite{cai2024medusa}. The answer is then the head's box unless the
verifier clearly disagrees, and is no longer guaranteed to equal $y^\ast$. We choose $\alpha\in\{0.7,0.5,0.3\}$ on
RefCOCO val only.

\subsection{Training one model for both roles}
\label{sec:method:train}

We train the box head jointly with the language model and vision merger, fully fine-tuned as in \jag (the ablation
also uses a LoRA adapter~\cite{hu2022lora} on the same modules). Each step prefills a batch of images once and runs, per image, up to four box branches and
the same number of verifier branches whose targets are the ground-truth boxes rendered as coordinate tokens. The loss
is the head loss plus the verifier's token cross-entropy with weight 1. Because the weights are shared, one image prefill serves both
roles at inference. Joint training is as accurate as training two separate models (\cref{sec:results:ablations}).

\paragraph{Dual mode.} The trained weights serve box and verifier branches and the image prefill; any other question
about the same image is answered by the base weights (an unloaded adapter, or a kept base copy after full fine-tuning),
so its answer is exactly the base model's. The cost is a second, base-model prefill when a request mixes box and
non-box questions, and for full fine-tuning a second copy of the weights. An activated LoRA (aLoRA)~\cite{greenewald2025alora} avoids
the second prefill: the adapter is active only from the box-query token onwards, so the image prefill is the base
model's and is shared by every kind of question. The price is that the image states are never adapted, which costs
grounding accuracy, most of all at high IoU (\cref{sec:results:ablations}). We use dual mode.

%% file: sections/setup.tex
\section{Experimental Setup}
\label{sec:setup}

\paragraph{Backbones.} Our main model uses Qwen3-VL-2B~\cite{qwen3vl}. For the equal-size comparison we use
Qwen3.5-0.8B, \jag's own backbone, a hybrid of Gated DeltaNet linear-attention and full-attention layers.

\paragraph{Training data.} We train on the RefCOCO, RefCOCO+ and RefCOCOg~\cite{kazemzadeh2014referitgame,yu2016modeling,mao2016generation}
training splits after removing every image that appears in any evaluation split. The standard union of these training
sets shares images with the evaluation splits of the other datasets; we do not train on those images.
The RefCOCO-family splits are drawn per dataset, so \N{leak-frac}\% of the evaluation rows
(\N{leak-frac-min}--\N{leak-frac-max}\% per split) have a target object that occurs in the training split of another
dataset. Most published models, \jag included, train on that union. We report every comparison on all rows and, as fixed before any results, on the
\emph{clean} rows whose target object is in no training split, which no model in our tables has seen
(\cref{sec:results:clean}).

\paragraph{Training budget.} \method-0.8B and \method-2B fully fine-tune the language model and vision merger for 36k
steps of 4 images $\times$ 4 expressions, which is 2.00 epochs of our training set, the same budget as \jag.
The 2B LoRA variants of the ablation ($r=64$, optionally plus a LoRA on the vision merger) use 12k steps. A 54k-step (3-epoch) run is reported separately as \emph{more} budget than \jag.

\paragraph{Evaluation.} We report \accat{0.5}, \accat{0.75} and \accat{0.9} on the 8 val/test splits of RefCOCO,
RefCOCO+ and RefCOCOg, and their unweighted mean. $\tau$ and $\alpha$ are chosen on RefCOCO val and applied unchanged
to the other splits. \method sees each image at its $640^2$-pixel training budget and \jag at its own $512^2$; we
also run each at the other's resolution.

\paragraph{Outside RefCOCO.} Ref-L4 test~\cite{chen2024refl4} has \N{refl4-n} expressions with ground-truth boxes,
averaging \N{refl4-words} words. Its \N{refl4-ono} Objects365~\cite{shao2019objects365} rows are on images no model here was trained
on. Its COCO images all come from RefCOCO-family evaluation annotations, and \N{refl4-ntrain} rows are on images that
are also in a RefCOCO-family training split, which models trained on the union have seen; we report the \N{refl4-ncoco} COCO rows on the
other images separately. Every model runs exactly as in its RefCOCO evaluation, with no Ref-L4 tuning.

\paragraph{Baselines.} We run every baseline with its released weights in our harness on exactly our rows. \jag-0.8B
is the strongest open regression-head model; its 2B model is not released. We score the zero-shot coordinate-token answers of Qwen3-VL-2B and -4B with their native grounding prompt (greedy,
0--1000 coordinates); these were run on the first 3000 rows of RefCOCO val, testA and testB, and all systems are
re-scored on exactly those rows for that comparison. We run the released LocateAnything-3B~\cite{wang2026locateanything} on all
8 splits with its model-card single-instance prompt, hybrid parallel box decoding, greedy, bf16, batch 1 and native
image resolution; on RefCOCO testA it reaches \accat{0.5} \N{la-testa-a}, against the 93.23 reported by \jag. Hi-Token~\cite{zhu2026hitoken} (Qwen2.5-VL-3B with axis-specific digit tokens) and
NExT-Chat-7B~\cite{zhang2023nextchat} (a box decoder on an LLM embedding) run the same way, each with its released weights,
model-card prompt and greedy bf16 decoding at the resolution its processor uses; for NExT-Chat we use the README's
localisation prompt and read the box from its location decoder, without the mask head.

\paragraph{Latency.} One RTX A6000 GPU rented from RunPod, batch 1, $512^2$-pixel inputs with preprocessing on the GPU, on 256 RefCOCO testA
questions. \jag and \method are timed on the same 256 images (one question each), CUDA-graphed with all graphs
captured before timing; their graphed paths produce the same boxes and pass counts as eager execution. The
coordinate-token baselines run in vLLM~0.31 with CUDA graphs on the same 256 images: Qwen3-VL-2B writing its full JSON
answer, and, decoding only the coordinate tokens \method verifies, Qwen3-VL-2B and \method-2B's own weights given the
verifier's prompt with the JSON preamble already written. Qwen3-VL-2B in Hugging Face eager is timed on 256 of the
first 266 testA questions. Qwen3.5-0.8B (\jag's base model, with \jag's base prompt), LocateAnything-3B, Hi-Token
and NExT-Chat-7B are timed on the same 256 images as \jag, with the inference code and settings of their accuracy runs.
All latencies in this paper are from this one machine.

%% file: sections/results.tex
\section{Results}
\label{sec:results}

\begin{table*}[t]
\caption{\accat{0.5} against published grounding models. Rows above ``Ours'' are the numbers reported by each paper
(source tables are listed in our repository); several papers report only test splits, so the last column averages the
5 test splits, which every row reports. Our rows (tinted) are measured in our harness. Bold: best, underlined: second
best per column, among models with at most 3B parameters. Training data differ: most models train on the union of the RefCOCO-family training sets, which contains
\N{leak-frac}\% of the evaluation targets (\cref{sec:results:clean}); we exclude every evaluation image.}
\label{tab:literature}
\centering
\scriptsize
\setlength{\tabcolsep}{3.2pt}
\input{generated/literature}
\end{table*}

\subsection{Comparison with published models}
\label{sec:results:literature}

\Cref{tab:literature} places \method among published grounding models at \accat{0.5}, the one metric all of them
report. The one-pass \method-2B head averages \N{lit-locus2b-head-test} on the 5 test splits. That is above every
published model with at most 3B parameters except Florence-2-L~\cite{xiao2023florence2}
(\N{lit-florence2l-test}): LocateAnything-3B~\cite{wang2026locateanything} reaches \N{lit-locateanything3b-test},
\jag-0.8B \N{lit-jag08b-test} and Qwen2.5-VL-3B~\cite{qwen25vl} \N{lit-qwen25vl3b-test}. It is also above the 7B--13B
coordinate-token models Qwen2.5-VL-7B (\N{lit-qwen25vl7b-test}), Ferret-13B~\cite{you2023ferret}
(\N{lit-ferret13b-test}) and Shikra-7B~\cite{chen2023shikra} (\N{lit-shikra7b-test}), and above the specialist
Grounding DINO~\cite{liu2023groundingdino} (\N{lit-groundingdino-test}). Florence-2-L is pretrained on 126M images
with region-level annotations, and the 72B--78B generalists remain ahead. The largest differences between \method and
other models are at higher IoU thresholds, which this table does not report.

\begin{table}[t]
\caption{Grounding accuracy, mean over the 8 val and test splits of RefCOCO, RefCOCO+ and RefCOCOg, all run in our
harness on the same rows. Baselines use their released weights and model-card prompts at the resolution their
processor uses ($^\dagger$native; NExT-Chat a $336^2$ padded square; \jag $512^2$ pixels); \method runs at its $640^2$
training budget. \method-0.8B uses \jag's backbone and training budget. Passes: mean sequential branch passes per
question, including the greedy steps of the few questions that do not converge within 8 Jacobi passes. Our rows are
tinted.}
\label{tab:main}
\centering
\footnotesize
\setlength{\tabcolsep}{2.4pt}
\resizebox{\columnwidth}{!}{%
\begin{tabular}{llccc}
\toprule
Model & Output & \accat{0.5}\up & \accat{0.75}\up & \accat{0.9}\up \\
\midrule
\jag-0.8B & head & \N{jag-abs-a} & \N{jag-abs-b} & \N{jag-abs-c} \\
LocateAnything-3B$^\dagger$ & box tokens & \N{la-abs-a} & \N{la-abs-b} & \N{la-abs-c} \\
Hi-Token-3B$^\dagger$ & digits & \N{hitoken-abs-a} & \N{hitoken-abs-b} & \N{hitoken-abs-c} \\
NExT-Chat-7B & decoder & \N{nextchat-abs-a} & \N{nextchat-abs-b} & \N{nextchat-abs-c} \\
\midrule
\ours  & head (1 pass) & \N{twob-head-abs-a} & \N{twob-head-abs-b} & \N{twob-head-abs-c} \\
\ours  & exact (\N{twob-passes}) & \N{twob-verified-abs-a} & \N{twob-verified-abs-b} & \N{twob-verified-abs-c} \\
\ours \multirow{-3}{*}{\method-2B} & $\alpha=\N{twob-alpha}$ (\N{twob-typical-passes}) & \N{twob-typical-abs-a} & \N{twob-typical-abs-b} & \N{twob-typical-abs-c} \\
\midrule
\ours  & head (1 pass) & \N{small-head-abs-a} & \N{small-head-abs-b} & \N{small-head-abs-c} \\
\ours  & exact (\N{small-passes}) & \N{small-verified-abs-a} & \N{small-verified-abs-b} & \N{small-verified-abs-c} \\
\ours \multirow{-3}{*}{\method-0.8B} & $\alpha=\N{small-alpha}$ (\N{small-typical-passes}) & \N{small-typical-abs-a} & \N{small-typical-abs-b} & \N{small-typical-abs-c} \\
\bottomrule
\end{tabular}}
\end{table}

\subsection{Accuracy across IoU thresholds}
\label{sec:results:main}

\begin{figure}[t]
\centering
\includegraphics[width=\columnwidth]{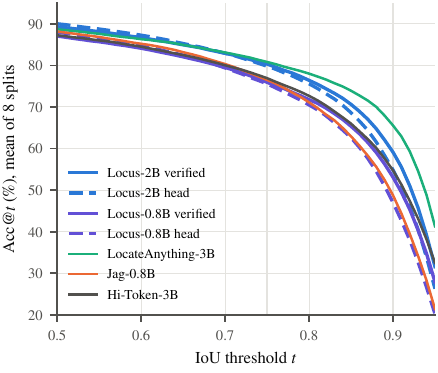}
\caption{Accuracy at each IoU threshold $t$, mean over the 8 splits, for the models of \cref{tab:main}. \method's
heads are dashed and its verified answers solid. The gap between a head and its verified answer opens above
$t\approx0.8$; LocateAnything-3B, trained on far more grounding data, is ahead at high thresholds.}
\label{fig:iou}
\end{figure}

\Cref{tab:main} compares \method with the open grounding models we ran on exactly our rows, and \cref{fig:iou} shows
the whole IoU range; \cref{fig:qualitative} shows examples. Verification adds most where a head is weakest. On the same weights, it raises \method-2B's
\accat{0.9} from \N{twob-head-abs-c} (head) to \N{twob-verified-abs-c} (exact), in \N{twob-passes} passes on average,
and \method-0.8B's from \N{small-head-abs-c} to \N{small-verified-abs-c}. Typical acceptance keeps most of that gain
with fewer passes. At \accat{0.5} the head is the more accurate output (\N{twob-head-abs-a} vs
\N{twob-verified-abs-a} at 2B). Against regression heads, \method-2B's exact answer is \N{twob-verified-dc} points
above \jag-0.8B, the strongest open one, at \accat{0.9}; part of that margin is the larger backbone, and
\cref{sec:results:equal} compares the two at equal size.

\paragraph{Same resolution.} \method runs at its $640^2$ training budget and \jag at $512^2$. The resolution does not
explain the gap: at $512^2$, \method-2B's exact answer reaches \accat{0.9} \N{twobpx-verified-abs-c} and \method-0.8B's
\N{smallpx-verified-abs-c}, against \N{jag-abs-c} for \jag; given $640^2$, \jag reaches \N{jagpx-abs-c}
(\accat{0.5} \N{jagpx-abs-a}).

\paragraph{Head and token errors.} Verification inherits the precision of coordinate tokens and the recall of the
head (\cref{tab:zeroshot}, which also reports Qwen3-VL-4B). Without any grounding fine-tuning, Qwen3-VL-2B's greedy boxes are already more precise than every head at \accat{0.9}
(\N{zs-qwenvl-c}, against \N{zs-jag-c} for \jag), but \N{zs-qwenvl-a} at \accat{0.5}, below every fine-tuned head.
The two output types fail in different ways: tokens miss the object more often, heads place its edges less exactly.
\method's verified answer combines the head's \accat{0.5} level (\N{zs-twobverified-a}) with token-level precision
(\N{zs-twobverified-c}). Verification returns exactly the fine-tuned model's greedy tokens, whose \accat{0.9} is only
modestly above the zero-shot base's; the head's contribution is a draft from which those tokens are reached in
\N{twob-passes} passes instead of \N{twob-tokens} decoding steps, and, with the agreement rule, its own
\accat{0.5}.

\paragraph{Combining head and verifier.} The head and the verifier are good at different things: the head is better at \accat{0.5}
(\N{twob-head-abs-a} vs \N{twob-verified-abs-a}), the verifier at \accat{0.9}. A one-parameter agreement rule
combines them: return the verified box if its IoU with the draft is at least $\tau$, and the draft otherwise, with
$\tau=\N{twob-tau}$ chosen on RefCOCO val. It reaches \N{twob-selected-abs-a} / \N{twob-selected-abs-c} at
\accat{0.5} / \accat{0.9}, keeping nearly all of the verifier's high-IoU gain and part of the head's \accat{0.5}
lead. A learned selector does no better (\cref{sec:results:ablations}).

\paragraph{LocateAnything.} LocateAnything-3B~\cite{wang2026locateanything}, which decodes its box tokens in parallel,
is more accurate than \method at high IoU. Run in our harness on the same rows, it matches \method-2B at \accat{0.5} (\N{la-abs-a} vs
\N{twob-verified-abs-a}) and is ahead at \accat{0.75} (\N{la-abs-b} vs \N{twob-verified-abs-b}) and \accat{0.9}
(\N{la-abs-c} vs \N{twob-verified-abs-c}). The gap holds on the clean rows (\accat{0.9}
\N{clean-la-abs-c} vs \N{clean-twob-verified-abs-c}). Given $512^2$ pixels, fewer than \method's $640^2$, it still reaches \accat{0.9} \N{lapx-abs-c} (native \N{la-abs-c}; \accat{0.5} \N{lapx-abs-a}).
It is larger and was trained on more than 138 million grounding samples~\cite{wang2026locateanything}, against our
RefCOCO-family training splits.

\paragraph{Beyond RefCOCO.} Verification helps on Ref-L4 (\cref{tab:refl4}) as it does on RefCOCO. On COCO images in
no training split, \method-2B leads \jag at \accat{0.9} (\N{refl4-twobverified-coco-c} vs \N{refl4-jag-coco-c}) and
matches the zero-shot base model (\N{refl4-qwenvl-coco-c}). On the Objects365 images, by contrast, every
RefCOCO-trained model falls far behind the zero-shot base model: \accat{0.5} is \N{refl4-twobverified-o365-a} for
\method-2B and \N{refl4-jag-o365-a} for \jag, against \N{refl4-qwenvl-o365-a}.

\paragraph{Other models on Ref-L4.} LocateAnything-3B, trained on far
more grounding data, is the best model on Objects365 (\N{refl4-la-o365-a}, and \N{refl4-la-o365q1-a} on the smallest
quarter of targets) and on COCO at \accat{0.9} (\N{refl4-la-coco-c}). Hi-Token, at its processor's native resolution,
fails on large images: on Objects365 its \accat{0.5} falls from \N{refl4-hitoken-o365smallimg-a} on images under
0.5\,MP to \N{refl4-hitoken-o365bigimg-a} on the \N{refl4-o365-nbig} rows above 4\,MP, where every other model stays
within ten points of its small-image accuracy.

\paragraph{Small targets.} The Objects365 gap is target size. The median Objects365 target covers \N{refl4-relarea-o365}\% of the image, against
\N{refl4-relarea-train}\% in our training data, where none of the \N{refl4-ntrainrows} targets is that small (1st
percentile \N{refl4-relarea-train-p1}\%). On the quarter of Objects365 rows with the smallest targets (at most
\N{refl4-o365-q1-max}\%), \accat{0.5} is \N{refl4-twobverified-o365q1-a} for \method-2B, \N{refl4-jag-o365q1-a} for
\jag and \N{refl4-qwenvl-o365q1-a} zero-shot. On the largest quarter (at least \N{refl4-o365-q4-min}\%) the three
are close (\N{refl4-twobverified-o365q4-a}, \N{refl4-jag-o365q4-a}, \N{refl4-qwenvl-o365q4-a}). Fine-tuning on RefCOCO
alone therefore costs small-object grounding, for heads and coordinate tokens alike. The head is hit hardest: \method-2B's
head reaches \N{refl4-twobhead-o365q1-a} on the smallest quarter, and verification recovers only a little of it.

\paragraph{Web photos.} On openly licensed web photos with expressions written before either model was run, \method-2B and \jag mostly agree;
\method's boxes are tighter on two of 12 expressions, they disagree on one, and both fail on an object part
(\cref{fig:wild}).

\begin{table}[t]
\caption{Ref-L4 test: Objects365 rows (images no model was trained on), COCO rows (only images outside every
RefCOCO/+/g training split), and all \N{refl4-n} rows. Each model runs as in its RefCOCO evaluation, with no Ref-L4
tuning. Columns are \accat{0.5} and \accat{0.9}. $^\dagger$Native resolution.}
\label{tab:refl4}
\centering
\footnotesize
\setlength{\tabcolsep}{1.9pt}
\input{generated/refl4}
\end{table}

\subsection{Comparison at equal size and budget}
\label{sec:results:equal}

\jag-0.8B is the closest regression-head model, and \method-0.8B uses its backbone and training budget. At
\accat{0.5}, the head of \method-0.8B ties
\jag (\N{small-head-da}; pooled McNemar $p=\N{mc-small-head-a-p}$, image-bootstrap 95\% CI \N{mc-small-head-a-ci}), and
verification with typical acceptance ($\alpha=\N{small-alpha}$) adds \N{small-typical-dc} points at \accat{0.9} (CI \N{mc-small-typical-c-ci}, $p=\N{mc-small-typical-c-p}$),
significant on every split. Verification costs \accat{0.5} at this size (\N{small-verified-da}), unlike at 2B: the
0.8B verifier is weaker than its own head on easy boxes. A third epoch, more budget than \jag, brings the head to
\N{smallthree-head-da} at \accat{0.5}, still a statistical tie ($p=\N{mc-smallthree-head-a-p}$). On all rows, the two
are therefore at parity at \accat{0.5}, and \method is ahead at high IoU. On the rows neither model was trained on, the
equal-budget head is ahead at \accat{0.5} as well (\cref{sec:results:clean}).

\subsection{Evaluation on clean rows}
\label{sec:results:clean}

\begin{figure}[t]
\centering
\includegraphics[width=\columnwidth]{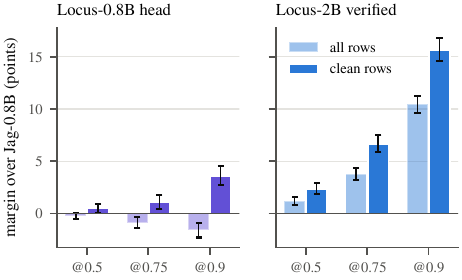}
\caption{Margin over \jag-0.8B on all rows and on clean rows, pooled over the 8 splits, with 95\% image-bootstrap
intervals. Every margin grows on the rows neither model was trained on.}
\label{fig:clean}
\end{figure}

Of the evaluation rows, \N{leak-frac}\% ask about an object that is in another RefCOCO-family dataset's training
split, so a model trained on the union of the three training sets, as most models in \cref{tab:literature} are, has
seen it under other expressions. Our training set excludes every evaluation image. On the remaining \emph{clean}
rows no model in our tables has seen the target. \Cref{fig:clean} (exact intervals in \cref{tab:clean}) repeats the
regression-head comparison of \cref{sec:results:equal} on them; \jag is trained on the union. It loses more on clean
rows than \method does (8-split mean
\accat{0.5} \N{jag-abs-a} $\rightarrow$ \N{clean-jag-abs-a}, against \N{twob-head-abs-a} $\rightarrow$
\N{clean-twob-head-abs-a} for the \method-2B head), and every margin grows. At equal size and budget the
\method-0.8B head is now ahead at \accat{0.5}, \N{mcclean-small-head-a-d} (CI \N{mcclean-small-head-a-ci},
McNemar $p=\N{mcclean-small-head-a-p}$), and the verified answers gain \N{mcclean-small-verified-c-d} points at
\accat{0.9}, against \N{mc-small-verified-c-d} on all rows. At 2B the verified \accat{0.9} gain is
\N{mcclean-twob-verified-c-d} points. Clean rows are harder for every model, including the untuned base model, so
the drop alone does not show memorisation. On clean rows, however, neither system has seen the target object
in training, so we consider this comparison the fairer one.

\subsection{Latency}
\label{sec:results:latency}

\begin{table}[t]
\caption{Latency and accuracy, all on one RTX A6000, batch 1, 256 RefCOCO testA questions, end to end from the image
file to the parsed box. Inputs are $512^2$ pixels except $^\dagger$native resolution and $^\ddagger$a $336^2$ padded
square, as each model's own processor requires. Engine: eager or CUDA graphs (PyTorch), HF (Hugging Face generate, eager), own (the authors' released inference
code, eager) or vLLM (CUDA graphs). ``Coord.\ only'': the
verifier's prompt with the JSON preamble already written, as in \method's verification; with \method-2B's weights the
greedy answer is \method-2B exact's, whose accuracy is shown (--: not run on the accuracy rows). Steps: sequential forward
passes after the image prefill (\method: the head branch plus verifier passes; token models: generated tokens; --:
not exposed by the model's generate). \accat{0.5} and \accat{0.9}: first 3000 rows of RefCOCO val, testA and testB, for \method from runs at the timed $512^2$ pixels. Bold: best,
underlined: second best; our rows are tinted.}
\label{tab:latency}
\centering
\footnotesize
\setlength{\tabcolsep}{3pt}
\input{generated/latency}
\end{table}

\begin{figure}[t]
\centering
\includegraphics[width=\columnwidth]{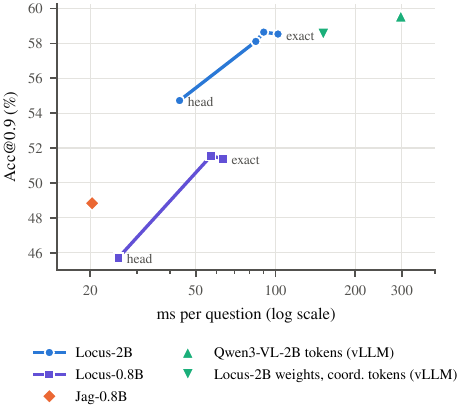}
\caption{\accat{0.9} against measured latency (RTX A6000, batch 1, 256 RefCOCO testA questions, every system with CUDA
graphs). Each \method line runs from the one-pass head through typical acceptance ($\alpha=0.5$, $0.7$; 2B only for
0.5) to exact verification. Accuracy is on the first 3000 rows of RefCOCO val, testA and testB, where the
coordinate-token baseline was run, with \method at the timed $512^2$ pixels. Qwen3-VL-2B decodes its full JSON answer
(\N{lat-qwenvllm-tokens} tokens on average) in vLLM.}
\label{fig:speed}
\end{figure}

In latency, \method sits between a regression head and a coordinate-token decoder (\cref{tab:latency,fig:speed}).
\jag, \method and the vLLM baselines share the same engineering: CUDA graphs and preprocessing on the GPU. A pure
regression head is faster: \jag-0.8B takes \N{lat-jag} ms and \method-0.8B's head \N{lat-small-head} ms, because
\method runs an image prefill followed by a question branch where \jag runs a single forward pass, and verification
multiplies that cost by the number of passes. Against
coordinate-token decoding the order reverses. The like-for-like reference is \method-2B's own weights served by vLLM,
given the verifier's prompt with the JSON preamble already written, so that it decodes exactly the
\N{lat-locusvllm-tokens} coordinate tokens \method verifies and returns the same greedy box: it takes
\N{lat-locusvllm} ms, and \method-2B exact takes \N{lat-twob-exact} ms, \N{lat-twob-exact-vscoords}$\times$ faster.
(The two engines round differently in bf16: at $640^2$ pixels \N{lat-locusvllm-same}\% of the boxes are
identical and \N{lat-locusvllm-five}\% within 5 units of the coordinate grid, the same kind of difference as between
two greedy runs in one engine, \cref{tab:exactness}.) Answer length matters more than the engine: the base
model decoding only its coordinates in vLLM takes \N{lat-qwenvllmcoords} ms, and writing its full JSON answer
(\N{lat-qwenvllm-tokens} tokens) \N{lat-qwenvllm} ms, \N{lat-twob-exact-vsvllm}$\times$ \method-2B exact. \method is thus
slower than a pure regression head and modestly faster than decoding the same coordinate tokens.

\paragraph{Released grounding models.} Measured on the same GPU, the released LocateAnything-3B, Hi-Token and
NExT-Chat-7B models take \N{lat-la}, \N{lat-hitoken} and \N{lat-nextchat} ms, against \N{lat-twob-head} ms for the
\method-2B head and \N{lat-twob-exact} ms for exact verification. They run
their authors' inference code without CUDA graphs, so with the same engineering each would be faster than measured
here. These numbers are therefore only indicative.

\begin{table}[tb]
\caption{Latency (ms, end to end, batch of one image) with $Q$ questions about the same image, on \N{latq-images}
RefCOCO testA images that have at least 8 expressions; same GPU and engineering as \cref{tab:latency}. \jag has no
shared prefill and is served the faster of $Q$ graphed requests and one batch of $Q$. Typical: $\alpha=\N{twob-alpha}$.}
\label{tab:latq}
\centering
\footnotesize
\setlength{\tabcolsep}{4pt}
\begin{tabular}{rcccc}
\toprule
& \jag-0.8B & \multicolumn{3}{c}{\method-2B} \\
\cmidrule(lr){3-5}
$Q$ & head & head & typical & exact \\
\midrule
1 & \N{latq-1-jag} & \N{latq-1-head} & \N{latq-1-typical} & \N{latq-1-exact} \\
2 & \N{latq-2-jag} & \N{latq-2-head} & \N{latq-2-typical} & \N{latq-2-exact} \\
4 & \N{latq-4-jag} & \N{latq-4-head} & \N{latq-4-typical} & \N{latq-4-exact} \\
8 & \N{latq-8-jag} & \N{latq-8-head} & \N{latq-8-typical} & \N{latq-8-exact} \\
\midrule
\multicolumn{5}{l}{\emph{per question at $Q=8$}} \\
8 & \N{latq-8-jag-perq} & \N{latq-8-head-perq} & \N{latq-8-typical-perq} & \N{latq-8-exact-perq} \\
\bottomrule
\end{tabular}
\end{table}

\paragraph{Several questions per image.} \method prefills the image once and runs every question as a branch, so its
head costs almost the same for one question as for eight (\N{latq-1-head} vs \N{latq-8-head}\,ms,
\cref{tab:latq}), \N{latq-8-head-perq}\,ms per question at $Q=8$ against \N{latq-8-jag-perq}\,ms for \jag, which
must encode the image with every question. \jag remains faster at $Q\le2$. Verification gains less: the questions of
one image are verified together, and the request waits for the slowest of them (\N{latq-8-exact-passes} sequential
passes at $Q=8$, including greedy steps after the cap of 8), so exact answers cost \N{latq-8-exact-perq}\,ms per
question at $Q=8$, still more than \jag's head.

\subsection{Ablations}
\label{sec:results:ablations}

\begin{table}[htb]
\caption{Ablations on adapter placement, adapted modules, and joint versus separate training. Mean over the 8 splits. All variants but \emph{separate} train the head and
the verifier jointly, as one model; tinted rows are the configurations we report as \method. $\ddagger$\,Evaluated on a different inference path; pass counts are only comparable within a path.}
\label{tab:ablation}
\centering
\footnotesize
\setlength{\tabcolsep}{1.8pt}
\begin{tabular}{lccccc}
\toprule
& \multicolumn{2}{c}{Head} & \multicolumn{2}{c}{Verified (exact)} & \\
\cmidrule(lr){2-3}\cmidrule(lr){4-5}
Variant & @0.5\up & @0.9\up & @0.5\up & @0.9\up & Passes\down \\
\midrule
\multicolumn{6}{l}{\emph{Qwen3.5-0.8B, 36k steps (2 epochs)}} \\
Joint aLoRA & \N{smalljoint-head-abs-a} & \N{smalljoint-head-abs-c} & \N{smalljoint-verified-abs-a} & \N{smalljoint-verified-abs-c} & \N{smalljoint-passes} \\
Dual LoRA & \N{smalllora-head-abs-a} & \N{smalllora-head-abs-c} & \N{smalllora-verified-abs-a} & \N{smalllora-verified-abs-c} & \N{smalllora-passes} \\
\quad + merger LoRA & \N{smallmerger-head-abs-a} & \N{smallmerger-head-abs-c} & \N{smallmerger-verified-abs-a} & \N{smallmerger-verified-abs-c} & \N{smallmerger-passes} \\
\ours Dual full FT (\method) & \N{small-head-abs-a} & \N{small-head-abs-c} & \N{small-verified-abs-a} & \N{small-verified-abs-c} & \N{small-passes} \\
\quad separate head, verifier & \N{smallsep-head-abs-a} & \N{smallsep-head-abs-c} & \N{smallsep-verified-abs-a} & \N{smallsep-verified-abs-c} & \N{smallsep-passes} \\
\quad 54k steps (3 epochs) & \N{smallthree-head-abs-a} & \N{smallthree-head-abs-c} & \N{smallthree-verified-abs-a} & \N{smallthree-verified-abs-c} & \N{smallthree-passes} \\
\midrule
\multicolumn{6}{l}{\emph{Qwen3-VL-2B, 12k steps unless stated}} \\
Joint aLoRA & \N{twobjoint-head-abs-a} & \N{twobjoint-head-abs-c} & \N{twobjoint-verified-abs-a} & \N{twobjoint-verified-abs-c} & --$^\ddagger$ \\
Dual LoRA & \N{twobdual-head-abs-a} & \N{twobdual-head-abs-c} & \N{twobdual-verified-abs-a} & \N{twobdual-verified-abs-c} & \N{twobdual-passes} \\
\quad + merger LoRA & \N{twoblora-head-abs-a} & \N{twoblora-head-abs-c} & \N{twoblora-verified-abs-a} & \N{twoblora-verified-abs-c} & \N{twoblora-passes} \\
\ours Dual full FT, 36k steps (\method) & \N{twob-head-abs-a} & \N{twob-head-abs-c} & \N{twob-verified-abs-a} & \N{twob-verified-abs-c} & \N{twob-passes} \\
\bottomrule
\end{tabular}
\end{table}

\paragraph{Dual mode vs aLoRA.} With an activated LoRA the image states stay those of the base model, and only the box
and verifier branches are adapted. This is cheap to serve but costs the head most of its precision: at 0.8B the aLoRA
head reaches \N{smalljoint-head-abs-c} \accat{0.9} against \N{smalllora-head-abs-c} in dual mode, and at 2B
\N{twobjoint-head-abs-c} against \N{twobdual-head-abs-c} (\cref{tab:ablation}). Adapting the image prefill matters for a
head that reads one token's state, much less for the verifier, which writes the box token by token: the verified
\accat{0.9} differs by at most a few points across all variants. At 0.8B the aLoRA draft also agrees less with the
verifier (\N{smalljoint-passes} vs \N{smalllora-passes} passes).

\paragraph{What to adapt.} A LoRA on the vision merger mainly sharpens the 2B head (\accat{0.9}
\N{twobdual-head-abs-c} $\rightarrow$ \N{twoblora-head-abs-c}) and leaves the verified answers almost unchanged; at 0.8B it
moves every number by less than half a point. Full fine-tuning of the language model and merger, as \jag does, gives the best 0.8B head and verifier
at \accat{0.9}. At 2B, full fine-tuning for 2 epochs lifts the head much further (\accat{0.9}
\N{twoblora-head-abs-c} $\rightarrow$ \N{twob-head-abs-c}) and cuts the passes from \N{twoblora-passes} to
\N{twob-passes}, while the verified \accat{0.9} moves little (\N{twoblora-verified-abs-c} vs \N{twob-verified-abs-c});
we use it as \method-2B. A third epoch improves the 0.8B head further (\N{smallthree-head-abs-a} / \N{smallthree-head-abs-c}) but
exceeds the 2-epoch budget of \cref{sec:results:equal}.

\paragraph{Separate head and verifier.} Training the head and the verifier as two models, each fully fine-tuned for 36k
steps (twice the training compute), does no better than one joint model: head \accat{0.5}
\N{smallsep-head-abs-a} vs \N{small-head-abs-a}, verified \accat{0.5} \N{smallsep-verified-abs-a} vs
\N{small-verified-abs-a} and \accat{0.9} \N{smallsep-verified-abs-c} vs \N{small-verified-abs-c}, at a similar
pass count (\N{smallsep-passes} vs \N{small-passes}; \cref{tab:ablation}). The pair also needs two image prefills and
two copies of the weights at inference.

\paragraph{Typical acceptance.} \Cref{fig:speed} and \cref{tab:alpha} sweep $\alpha$. Lowering it moves the answer towards the head:
\accat{0.5} rises towards the head's, \accat{0.9} falls towards it, and the pass count drops from
\N{twob-passes} to \N{twob-a03-passes}. $\alpha=0.7$, chosen on RefCOCO val, costs \accat{0.9} only
\N{twob-verified-abs-c} $\rightarrow$ \N{twob-a07-abs-c} for \N{twob-a07-passes} passes.

Two further ideas did not help, a learned head-versus-verifier selector and distilling the head on the verifier's own
greedy boxes (\cref{sec:appendix:negative}).

%% file: generated/literature.tex
\begin{tabular}{lll cccccccc cc}
\toprule
& & & \multicolumn{3}{c}{RefCOCO} & \multicolumn{3}{c}{RefCOCO+} & \multicolumn{2}{c}{RefCOCOg} & \multicolumn{2}{c}{Mean} \\
\cmidrule(lr){4-6}\cmidrule(lr){7-9}\cmidrule(lr){10-11}\cmidrule(lr){12-13}
Model & Params & Output & val & testA & testB & val & testA & testB & val & test & all 8 & 5 test \\
\midrule
\multicolumn{13}{l}{\emph{Specialist grounding models}} \\
MDETR~\cite{kamath2021mdetr} & -- & detector & 86.8 & 89.6 & 81.4 & 79.5 & 84.1 & 70.6 & 81.6 & 80.9 & 81.8 & 81.3 \\
PolyFormer~\cite{liu2023polyformer} & -- & polygon tokens & 90.4 & 92.9 & 87.2 & 85.0 & 89.8 & 78.0 & 85.8 & 85.9 & 86.9 & 86.8 \\
UNINEXT~\cite{yan2023uninext} & 0.74B & detector & 92.6 & \second{94.3} & \second{91.5} & 85.2 & 89.6 & 79.8 & \second{88.7} & 89.4 & 88.9 & 88.9 \\
Grounding DINO~\cite{liu2023groundingdino} & -- & detector & 90.6 & 93.2 & 88.2 & 88.2 & 89.0 & 75.9 & 86.1 & 87.0 & 87.3 & 86.7 \\
\midrule
\multicolumn{13}{l}{\emph{Vision-language models (reported)}} \\
Shikra-7B~\cite{chen2023shikra} & 7B & tokens & 87.0 & 90.6 & 80.2 & 81.6 & 87.4 & 72.1 & 82.3 & 82.2 & 82.9 & 82.5 \\
Ferret-13B~\cite{you2023ferret} & 13B & tokens & 89.5 & 92.4 & 84.4 & 82.8 & 88.1 & 75.2 & 85.8 & 86.3 & 85.6 & 85.3 \\
NExT-Chat-7B~\cite{zhang2023nextchat} & 7B & embedding + decoder & -- & 89.7 & 77.0 & -- & 83.8 & 66.2 & -- & 79.3 & -- & 79.2 \\
Florence-2-L~\cite{xiao2023florence2} & 0.77B & tokens & \best{93.4} & \best{95.3} & \best{92.0} & \best{88.3} & \best{92.9} & \best{83.6} & \best{91.2} & \best{91.7} & \best{91.0} & \best{91.1} \\
Qwen2.5-VL-3B~\cite{qwen25vl} & 3B & tokens & 89.1 & 91.7 & 84.0 & 82.4 & 88.0 & 74.1 & 85.2 & 85.7 & 85.0 & 84.7 \\
Qwen2.5-VL-7B~\cite{qwen25vl} & 7B & tokens & 90.0 & 92.5 & 85.4 & 84.2 & 89.1 & 76.9 & 87.2 & 87.2 & 86.6 & 86.2 \\
Qwen2.5-VL-72B~\cite{qwen25vl} & 72B & tokens & 92.7 & 94.6 & 89.7 & 88.9 & 92.2 & 83.7 & 89.9 & 90.3 & 90.2 & \second{90.1} \\
InternVL2.5-78B~\cite{chen2024internvl25} & 78B & tokens & 93.7 & 95.6 & 92.5 & 90.4 & 94.7 & 86.9 & 92.7 & 92.2 & 92.3 & 92.4 \\
LocateAnything-3B~\cite{wang2026locateanything} & 3B & parallel box tokens & -- & 93.2 & 89.3 & -- & 88.0 & 79.6 & -- & 88.5 & -- & 87.7 \\
Jag-0.8B~\cite{zhu2026jag} & 0.8B & regression head & -- & 93.9 & 88.4 & -- & 90.7 & 80.1 & -- & 87.7 & -- & 88.2 \\
\midrule
\multicolumn{13}{l}{\emph{Ours (measured)}} \\
\ours \method-0.8B & 0.8B & head & 91.0 & 93.8 & 88.0 & 85.6 & 90.6 & 80.7 & 86.9 & 87.0 & 88.0 & 88.0 \\
\ours \method-2B & 2B & head & \second{93.1} & \second{94.3} & 90.7 & \second{87.9} & \second{91.9} & \best{83.6} & \second{88.7} & \second{89.9} & \second{90.0} & \second{90.1} \\
\ours \method-2B & 2B & head + tokens (rule) & 92.5 & 94.0 & 90.0 & 87.6 & 91.7 & \second{83.4} & 88.0 & 89.6 & 89.6 & 89.7 \\
\ours \method-2B & 2B & tokens (verified) & 92.1 & 93.9 & 89.5 & 87.3 & 91.5 & 82.8 & 87.8 & 89.3 & 89.3 & 89.4 \\
\bottomrule
\end{tabular}

%% file: generated/refl4.tex
\begin{tabular}{lcccccc}
\toprule
 & \multicolumn{2}{c}{Objects365} & \multicolumn{2}{c}{COCO} & \multicolumn{2}{c}{All} \\
\cmidrule(lr){2-3}\cmidrule(lr){4-5}\cmidrule(lr){6-7}
Model & @0.5\up & @0.9\up & @0.5\up & @0.9\up & @0.5\up & @0.9\up \\
\midrule
\jag-0.8B & 35.8 & 7.4 & 89.2 & 40.0 & 70.0 & 33.9 \\
Qwen3-VL-2B (0-shot) & \second{74.5} & \second{32.7} & 88.6 & \second{57.1} & \second{83.9} & \second{50.2} \\
Hi-Token-3B$^\dagger$ & 24.7 & 1.1 & 88.0 & 44.9 & 65.0 & 35.9 \\
LocateAnything-3B$^\dagger$ & \best{78.0} & \best{49.5} & 90.2 & \best{64.4} & \best{86.2} & \best{60.5} \\
\midrule
\ours \method-0.8B head & 39.3 & 9.2 & 89.6 & 43.7 & 71.2 & 32.9 \\
\ours \method-0.8B verified & 42.2 & 16.7 & 88.7 & 50.5 & 71.8 & 39.6 \\
\ours \method-2B head & 40.0 & 11.2 & \best{91.9} & 52.4 & 72.6 & 38.7 \\
\ours \method-2B verified & 47.8 & 18.1 & \second{90.8} & \second{57.1} & 75.0 & 44.0 \\
\bottomrule
\end{tabular}

%% file: generated/latency.tex
\begin{tabular}{llrrcc}
\toprule
System & Engine & ms\down & Steps & @0.5\up & @0.9\up \\
\midrule
\multicolumn{6}{l}{\emph{One RTX A6000, batch 1, end to end}} \\
\jag-0.8B & eager & 53.1 & 0.0 & 91.5 & 48.8 \\
\jag-0.8B & graphs & \best{20.3} & 0.0 & 91.5 & 48.8 \\
Qwen3-VL-2B tokens & HF & 1484.6 & 41.8 & 88.5 & \second{59.6} \\
Qwen3-VL-2B tokens & vLLM & 297.2 & 41.8 & 88.5 & \second{59.6} \\
\quad coord.\ only & vLLM & 150.9 & 18.3 & -- & -- \\
\method-2B, coord.\ only & vLLM & 151.6 & 18.3 & 91.6 & 58.5 \\
Qwen3.5-0.8B tokens & HF & 1243.2 & 46.7 & 79.8 & 42.9 \\
Hi-Token-3B$^\dagger$ & HF & 577.7 & 18.0 & 90.3 & 53.6 \\
LocateAnything-3B$^\dagger$ & own & 347.5 & -- & 91.5 & \best{66.9} \\
NExT-Chat-7B$^\ddagger$ & own & 436.7 & -- & 84.1 & 12.4 \\
\midrule
\ours \method-0.8B head & graphs & \second{25.5} & 1.0 & 90.9 & 45.7 \\
\ours \method-0.8B $\alpha$=0.7 & graphs & 57.1 & 5.1 & 89.8 & 51.5 \\
\ours \method-0.8B exact & graphs & 63.5 & 6.1 & 89.7 & 51.4 \\
\ours \method-2B head & graphs & 43.5 & 1.0 & \best{92.4} & 54.7 \\
\ours \method-2B $\alpha$=0.5 & graphs & 84.3 & 4.4 & \second{91.8} & 58.1 \\
\ours \method-2B $\alpha$=0.7 & graphs & 90.3 & 4.9 & 91.7 & 58.6 \\
\ours \method-2B exact & graphs & 102.2 & 5.9 & 91.6 & 58.5 \\
\bottomrule
\end{tabular}

%% file: sections/limitations.tex
\section{Limitations and Conclusion}
\label{sec:limitations}

\paragraph{Limitations.} \method is slower than a pure regression head with equal engineering, and verification adds
passes in proportion to how often the draft is wrong. At 0.8B the verifier lowers \accat{0.5}, so the method's benefit
at that size is high-IoU accuracy only. Its boxes are at most as tight as the model's own coordinate tokens, and
LocateAnything-3B, larger and trained on more data, remains ahead at high IoU even at $512^2$ pixels. Exactness holds for the adapted model's
greedy decode, not for any other model, and typical acceptance gives it up. In bf16 it holds only up to rounding: most
verified boxes differ from a greedy run by a unit of the coordinate grid, as greedy runs with different batch shapes
differ from each other. Dual mode needs a second prefill when one request mixes box and non-box questions on the same
image. We train on RefCOCO-family data only. On Ref-L4's Objects365 images, where half the targets are smaller
than every RefCOCO-family training target, \method-2B reaches \accat{0.5} \N{refl4-twobverified-o365-a} where its zero-shot base model
reaches \N{refl4-qwenvl-o365-a} (\cref{tab:refl4}); \jag drops the same way. Adding training data with small targets
may close this gap; we have not tested it. Our backbones stop at 2B.

\paragraph{Conclusion.} A regression head and coordinate-token decoding fail in different ways: the head usually finds
the object, and the tokens place its edges. \method treats the head as a speculative drafter for the model's own
coordinate tokens, so the answer is the model's greedy box, reached in \N{twob-passes} passes, each checking every token in parallel, instead of
\N{twob-tokens} decoding steps. This recovers the high-IoU accuracy that regression heads give up
(\accat{0.9} \N{twob-head-abs-c} $\rightarrow$ \N{twob-verified-abs-c} at 2B on the same weights, and
\N{small-typical-dc} points over \jag at its size and budget, with typical acceptance), and
the answer is exact up to the rounding that affects greedy decoding equally. The same recipe could apply wherever a cheap head can draft a short structured answer that a VLM would otherwise
decode token by token, such as points, keypoints or several boxes per expression; we leave these, and larger
backbones, to future work.

%% file: sections/appendix.tex
\section{Supplementary material}
\label{sec:appendix}

\begin{figure*}[t]
\centering
\includegraphics[width=\textwidth]{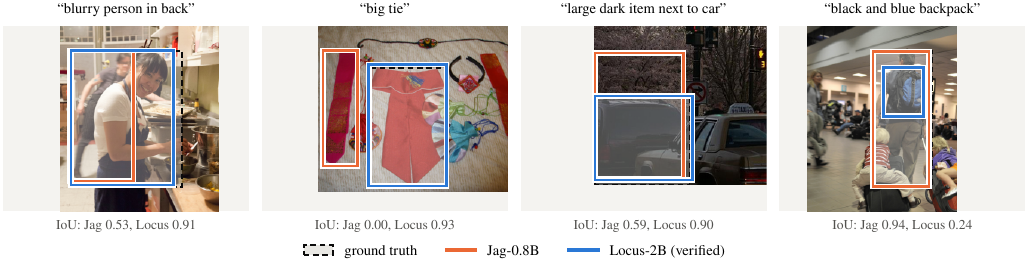}
\caption{\jag-0.8B and \method-2B (verified) on RefCOCO and RefCOCO+ test questions. The first three columns are the
first questions in split order where \method's IoU exceeds 0.9 and \jag's is below 0.6; the last is the first where
\method fails (IoU $<0.4$) and \jag succeeds (IoU $>0.8$). To skip typos and slang in the crowd-written expressions,
only expressions whose every word occurs at least 100 times in the training expressions are eligible. Crops are
centred on the three boxes.}
\label{fig:qualitative}
\end{figure*}

\begin{figure*}[t]
\centering
\includegraphics[width=\textwidth]{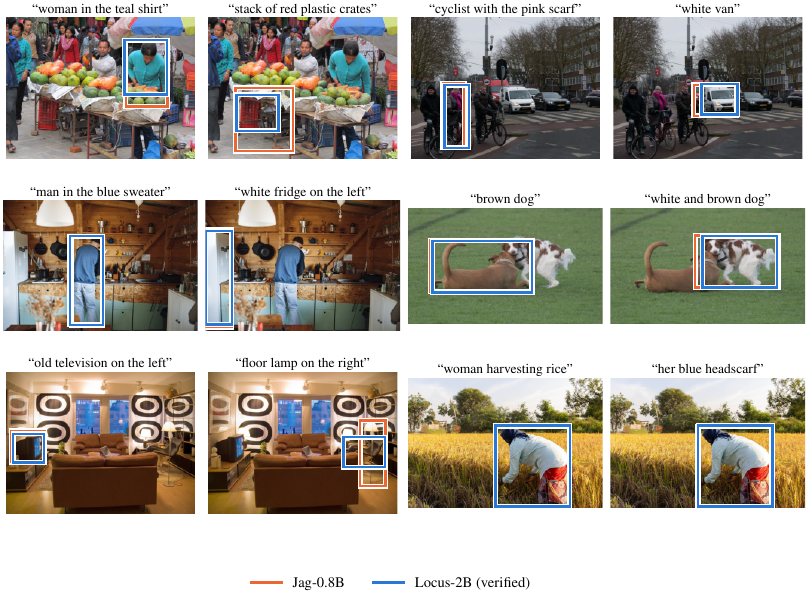}
\caption{\jag-0.8B and \method-2B (verified) on openly licensed photos from Wikimedia Commons, at the
same $512^2$-pixel budget. The 12 expressions were written before either model was run, and every result is shown;
there is no ground truth, so we make no accuracy claim. The two models mostly agree; \method's boxes are tighter for
``woman in the teal shirt'' and ``stack of red plastic crates'', they disagree on ``floor lamp on the right'', and both
return the whole person for ``her blue headscarf''. Images (left to right, top to bottom, two expressions each):
\emph{Fruit Market, Old Kathmandu} by Sundar1 (CC BY-SA 3.0); Amsterdam cyclists by Fons Heijnsbroek (CC0);
\emph{Person cooks in a cozy wooden kitchen} by Shixart1985 (CC BY 2.0); \emph{Dogs playing 001} by Sardaka
(CC BY-SA 4.0); \emph{Living room with 1960s objects} by Kotivalo (CC BY-SA 4.0); \emph{Farmer manually harvesting
rice in Raichur, Karnataka} by Nanditha Gogate, WELL Labs (CC BY-SA 4.0).}
\label{fig:wild}
\end{figure*}

\subsection{Additional tables}
\label{sec:appendix:tables}

\begin{table}[H]
\caption{Coordinate-token decoding without grounding fine-tuning, on the first 3000 rows of RefCOCO val, testA and
testB (mean of the three), with every system scored on exactly these rows. The Qwen models' pretraining includes
RefCOCO training data.}
\label{tab:zeroshot}
\centering
\footnotesize
\setlength{\tabcolsep}{2.2pt}
\begin{tabular}{llccc}
\toprule
Model & Output & \accat{0.5}\up & \accat{0.75}\up & \accat{0.9}\up \\
\midrule
Qwen3-VL-2B, zero-shot & tokens & \N{zs-qwenvl-a} & \N{zs-qwenvl-b} & \N{zs-qwenvl-c} \\
Qwen3-VL-4B, zero-shot & tokens & \N{zs-qwenvlfour-a} & \N{zs-qwenvlfour-b} & \N{zs-qwenvlfour-c} \\
\jag-0.8B & head & \N{zs-jag-a} & \N{zs-jag-b} & \N{zs-jag-c} \\
\midrule
\ours \method-0.8B & head & \N{zs-smallhead-a} & \N{zs-smallhead-b} & \N{zs-smallhead-c} \\
\ours  & verified & \N{zs-smallverified-a} & \N{zs-smallverified-b} & \N{zs-smallverified-c} \\
\ours \method-2B & head & \N{zs-twobhead-a} & \N{zs-twobhead-b} & \N{zs-twobhead-c} \\
\ours  & verified & \N{zs-twobverified-a} & \N{zs-twobverified-b} & \N{zs-twobverified-c} \\
\bottomrule
\end{tabular}
\end{table}

\begin{table}[H]
\caption{Exactness in floating point (\method-2B, RefCOCO val; the 505-row runs are its first questions). Differ: \%
of boxes with any coordinate different; $\ge$10: \% whose largest coordinate change of at least 10 on the 0--1000 grid.
Passes: mean verifier passes, against \N{exact-graphedfull-greedy-passes} greedy steps.}
\label{tab:exactness}
\centering
\footnotesize
\setlength{\tabcolsep}{4pt}
\begin{tabular}{llrrrr}
\toprule
Compared & Path & Rows & Differ & $\ge$10 & Passes \\
\midrule
ver.\ vs.\ greedy & graphed, bf16 & \N{exact-graphedfull-n} & \N{exact-graphedfull-differ} & \N{exact-graphedfull-gapten} & \N{exact-graphedfull-other-passes} \\
ver.\ vs.\ greedy & eager, bf16 & \N{exact-eagerbf16-n} & \N{exact-eagerbf16-differ} & \N{exact-eagerbf16-gapten} & \N{exact-eagerbf16-other-passes} \\
ver.\ vs.\ greedy & eager, fp32 & \N{exact-eagerfp32-n} & \N{exact-eagerfp32-differ} & \N{exact-eagerfp32-gapten} & \N{exact-eagerfp32-other-passes} \\
\midrule
greedy vs.\ greedy & batch 2/1 & \N{exact-greedybatch-n} & \N{exact-greedybatch-differ} & \N{exact-greedybatch-gapten} & -- \\
greedy vs.\ greedy & eager/graphed & \N{exact-greedygraph-n} & \N{exact-greedygraph-differ} & \N{exact-greedygraph-gapten} & -- \\
greedy vs.\ greedy & bf16/fp32 & \N{exact-greedyprecision-n} & \N{exact-greedyprecision-differ} & \N{exact-greedyprecision-gapten} & -- \\
\bottomrule
\end{tabular}
\end{table}

\begin{table}[H]
\caption{The comparison restricted to evaluation rows whose target object is in no RefCOCO-family training split
(\N{leak-frac}\% of rows are excluded). Pooled over the 8 splits, row-weighted; CI from a bootstrap over images.}
\label{tab:clean}
\centering
\footnotesize
\setlength{\tabcolsep}{4pt}
\resizebox{\columnwidth}{!}{%
\begin{tabular}{llccc}
\toprule
& & \multicolumn{3}{c}{$\Delta$ vs \jag-0.8B, clean rows (95\% CI)} \\
\cmidrule(lr){3-5}
Model & Output & \accat{0.5} & \accat{0.75} & \accat{0.9} \\
\midrule
\method-2B & head & \N{mcclean-twob-head-a-d} & \N{mcclean-twob-head-b-d} & \N{mcclean-twob-head-c-d} \\
 & & \N{mcclean-twob-head-a-ci} & \N{mcclean-twob-head-b-ci} & \N{mcclean-twob-head-c-ci} \\
 & verified & \N{mcclean-twob-verified-a-d} & \N{mcclean-twob-verified-b-d} & \N{mcclean-twob-verified-c-d} \\
 & & \N{mcclean-twob-verified-a-ci} & \N{mcclean-twob-verified-b-ci} & \N{mcclean-twob-verified-c-ci} \\
\midrule
\method-0.8B & head & \N{mcclean-small-head-a-d} & \N{mcclean-small-head-b-d} & \N{mcclean-small-head-c-d} \\
 & & \N{mcclean-small-head-a-ci} & \N{mcclean-small-head-b-ci} & \N{mcclean-small-head-c-ci} \\
 & typical & \N{mcclean-small-typical-a-d} & \N{mcclean-small-typical-b-d} & \N{mcclean-small-typical-c-d} \\
 & & \N{mcclean-small-typical-a-ci} & \N{mcclean-small-typical-b-ci} & \N{mcclean-small-typical-c-ci} \\
\bottomrule
\end{tabular}}
\end{table}

\begin{table}[H]
\caption{Effect of the typical-acceptance threshold $\alpha$ (\method-2B, mean over the 8 splits).}
\label{tab:alpha}
\centering
\footnotesize
\setlength{\tabcolsep}{5pt}
\begin{tabular}{lcccc}
\toprule
Acceptance & \accat{0.5}\up & \accat{0.75}\up & \accat{0.9}\up & Passes\down \\
\midrule
exact ($\alpha=1$) & \N{twob-verified-abs-a} & \N{twob-verified-abs-b} & \N{twob-verified-abs-c} & \N{twob-passes} \\
$\alpha=0.7$ & \N{twob-a07-abs-a} & \N{twob-a07-abs-b} & \N{twob-a07-abs-c} & \N{twob-a07-passes} \\
$\alpha=0.5$ & \N{twob-a05-abs-a} & \N{twob-a05-abs-b} & \N{twob-a05-abs-c} & \N{twob-a05-passes} \\
$\alpha=0.3$ & \N{twob-a03-abs-a} & \N{twob-a03-abs-b} & \N{twob-a03-abs-c} & \N{twob-a03-passes} \\
head only & \N{twob-head-abs-a} & \N{twob-head-abs-b} & \N{twob-head-abs-c} & 0 \\
\bottomrule
\end{tabular}
\end{table}

\subsection{Negative results}
\label{sec:appendix:negative}

\paragraph{Learned selector.} The agreement rule uses one feature. We also trained a small MLP on RefCOCO
val to choose between the draft and the verified box from the head confidence, the pass count, the draft--verified IoU
and the box geometry (5 seeds). On the 7 held-out splits it matches the rule at 2B (\N{selector-twob-sel-a} /
\N{selector-twob-sel-c} vs \N{selector-twob-rule-a} / \N{selector-twob-rule-c} at \accat{0.5} / \accat{0.9}). At 0.8B
it gains \accat{0.5} over the rule (\N{selector-small-sel-a} vs \N{selector-small-rule-a}) but stays below \jag
(\N{selector-small-jag-a}). Which of the two boxes is better is hard to predict from these signals, so we keep the rule.

\paragraph{Drafter distillation.} Training the 2B head on the verifier's own greedy boxes instead of the
ground truth should make drafts easier to accept. On RefCOCO val it changes the pass count only from
\N{distill-parent-passes} to \N{distill-distill-passes}, with the same accuracy
(\N{distill-distill-abs-a} / \N{distill-distill-abs-c} vs \N{distill-parent-abs-a} / \N{distill-parent-abs-c}). A
regression head rarely lands on the verifier's exact integers: across the 8 splits, \N{twob-draft-exact4}\% of
\method-2B drafts match all four greedy integers and \N{twob-draft-exact2}\% match at least two, so nearly every
question needs a correction pass whatever the head is trained on.